\pdfoutput=1
\documentclass[11pt]{article}

\usepackage[]{ACL2023}
\usepackage{graphicx}
\usepackage{pifont}
\usepackage{acronym}
\usepackage{times}
\usepackage{booktabs}
\usepackage{adjustbox}
\usepackage{latexsym}
\usepackage{xspace}
\usepackage{transparent}

\usepackage[T1]{fontenc}
\usepackage{multirow}
\usepackage{CJKutf8}
\usepackage{subcaption}

\usepackage[utf8]{inputenc}

\usepackage{microtype}

\usepackage{inconsolata}
\usepackage{colortbl}
\newcommand{\smallincreasebg}[1]{{\cellcolor[HTML]{bdbdbd}{#1}}} %
\newcommand{\tinyincreasebg}[1]{{\cellcolor[HTML]{e5e4e2}{#1}}} %e5e4e2
\definecolor{mygray}{gray}{.9}

\newcommand{\header}[1]{\vspace*{1mm}\noindent\textbf{#1}.}

\newcommand{\ie}{\textit{i.e., }}

\newcommand{\OurData}{McMarket\xspace}

\newcommand{\OurDatarank}{McMarket${_q}$\xspace}
\newcommand{\OurDataans}{McMarket${_r}$\xspace}
\acrodef{MCPQA}{multilingual cross-market product-based question answering}
\acrodef{AG}{review-based answer generation}
\acrodef{QR}{product-related question ranking}
\acrodef{LLMs}{large language models}
\acrodef{PQA}{Product-related question answering}
\title{Unlocking Markets: A Multilingual Benchmark to Cross-Market Question Answering}

\author{Yifei Yuan$^{1}$, Yang Deng$^{2}$, Anders Søgaard$^{1}$, Mohammad Aliannejadi$^{3}$ \\
  $^1$University of Copenhagen, $^2$Singapore Management University, $^3$University of Amsterdam \\
  \texttt{\{yiya,soegaard\}@di.ku.dk} \\
  \texttt{ydeng@smu.edu.sg, m.aliannejadi@uva.nl }}

\begin{document}
\maketitle
\begin{abstract}
Users post numerous product-related questions on e-commerce platforms, affecting their purchase decisions. Product-related question answering (PQA) entails utilizing product-related resources to provide precise responses to users. We propose a novel task of Multilingual Cross-market Product-based Question Answering (MCPQA) and define the task as providing answers to product-related questions in a main marketplace by utilizing information from another resource-rich auxiliary marketplace in a multilingual context. We introduce a large-scale dataset comprising over 7 million questions from 17 marketplaces across 11 languages. We then perform automatic translation on the Electronics category of our dataset, naming it as \OurData. We focus on two subtasks: \acl{AG} and \acl{QR}. For each subtask, we label a subset of \OurData using an LLM and further evaluate the quality of the annotations via human assessment. We then conduct experiments to benchmark our dataset, using models ranging from traditional lexical models to LLMs in both single-market and cross-market scenarios across \OurData and the corresponding LLM subset. Results show that incorporating cross-market information significantly enhances performance in both tasks.
\end{abstract}
% Answers are obtained either by generating or ranking from product-related resources (e.g., reviews, questions). 

\section{Introduction}
Online shoppers on platforms such as Amazon post numerous questions related to specific products every day~\cite{McAuley2015AddressingCA}. 
Product-related question answering (PQA) involves providing accurate and informative responses to these questions. By leveraging product-related information, such as reviews and product meta information, responses to product-related questions can be expanded, offering enhanced depth and authenticity for potential customers~\cite{Gupta2019AmazonQAAR}.
% Reviews play a crucial role in answering these questions by providing firsthand experiences from users who have interacted with the product.

\begin{figure}
\setlength{\abovecaptionskip}{5pt}   
\setlength{\belowcaptionskip}{0pt}
    \centering
    \includegraphics[width=\linewidth]{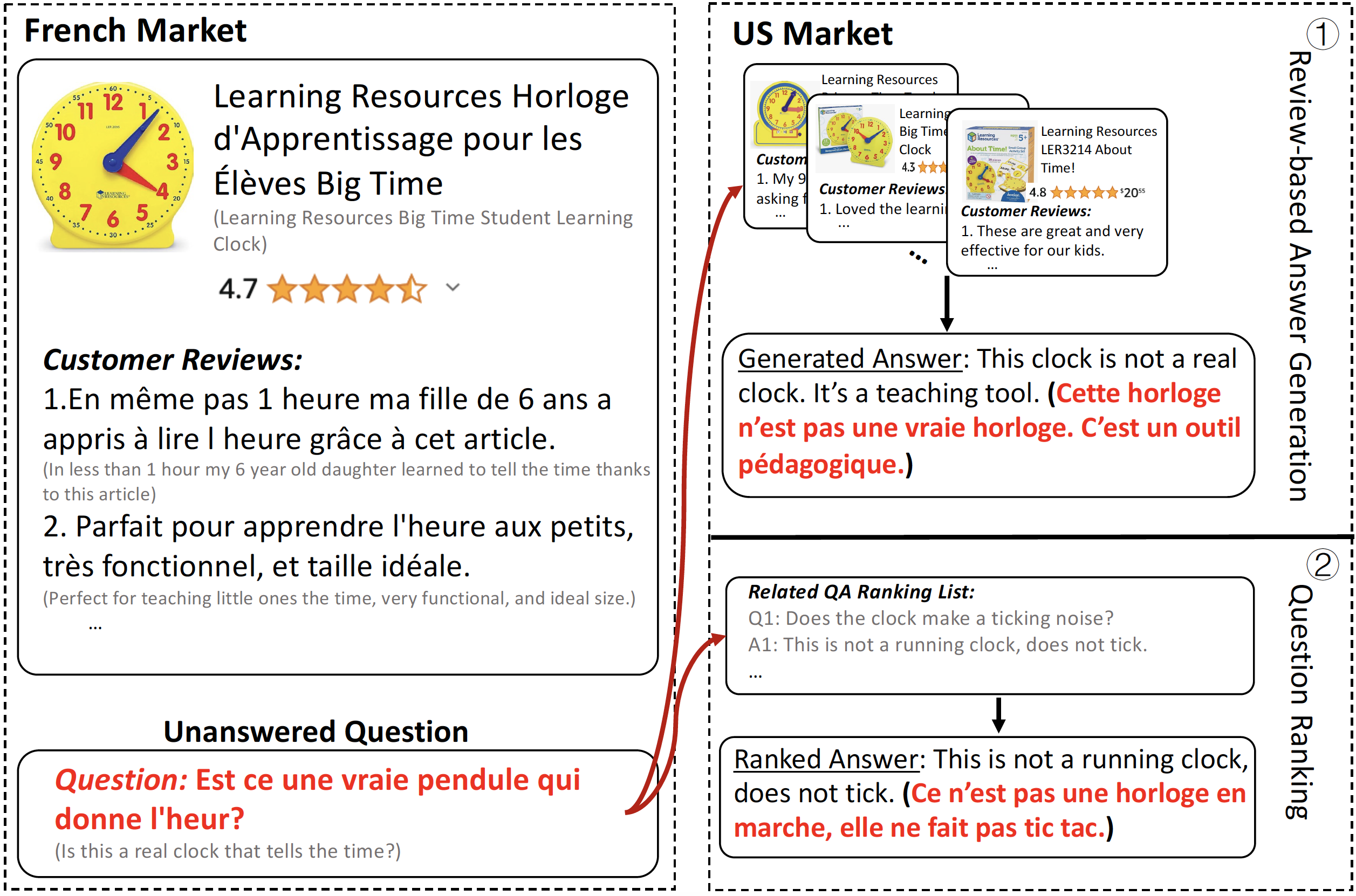}
    \caption{An example of enhancing product-related QA using cross-market data. \ding{172} depicts generating answers with cross-market reviews. \ding{173} depicts ranking-related cross-market questions to find the answer. }
    \label{fig:fig1}
    \vspace{-3mm}
\end{figure}

The recent success in cross-market \acs{PQA} underscores the capability to effectively leverage relevant questions from a resource-rich marketplace to address questions in a resource-scarce marketplace~\cite{Shen2023xPQACP,Ghasemi2023CrossMarketPQ}. In this work, we extend the hypothesis that leveraging knowledge from popular marketplaces can also enhance the quality of answers in less common marketplaces, even in a different language. As shown in Figure~\ref{fig:fig1}, for a question to a product in the French marketplace (denoted as \textbf{main marketplace}) asking if the clock is a real one, we can either address it by examining reviews of the same product or similar ones in the much larger US marketplace (denoted as \textbf{auxiliary marketplace}), or ranking related questions from both main and auxiliary marketplaces to find the answer. These multilingual reviews and related questions serve as valuable hints, by saying ``it's not a real clock.'', thereby providing crucial information for the pertinent question at hand. 

We, therefore, propose a novel task of \textit{Multilingual Cross-market Product-based Question Answering} (MCPQA). We define this task as producing the answer to a product-related question in an original marketplace, using information sourced from an auxiliary marketplace with richer resources,  within a multilingual setting. To this end, our initial goal is to address the following research question \textbf{RQ1}: \textit{In a multilingual context, how can we utilize an auxiliary marketplace to enhance question-answering in the main marketplace by leveraging product-related resources (\ie questions, reviews)?} To answer \textbf{RQ1}, we propose the first large-scale MCPQA dataset, covering 17 different marketplaces (including the \textbf{us} auxiliary marketplace and 16 main marketplaces) across 11 different languages from real Amazon product QA sources. Specifically, our dataset consists of over 7 million product-related questions with a total of 52 million product reviews. Different from existing \acs{PQA} datasets, more diverse information is provided in the dataset, exploring the possible answers with both questions and reviews available.  Additionally, we perform automatic translation on the Electronics category of the dataset, naming it \OurData. We then perform comprehensive data analysis on \OurData to address \textbf{RQ1}. We demonstrate a notable increase in the percentage of review-answerable questions across all marketplaces, with support from the auxiliary \textbf{us} marketplace. %, named \OurData, Specifically, we focus on two different tasks named  \textbf{\ac{AG}} and \textbf{\ac{QR}}.  compare different marketplaces in terms of the number of available questions over time and aiming to ascertain whether the product-related resources from the auxiliary marketplace can be effectively utilized to answer questions from other marketplaces on a smaller scale.

Given the recent success of \ac{LLMs} in NLP tasks~\cite{Touvron2023LLaMAOA,Achiam2023GPT4TR}, their potential application to the MCPQA task prompts our second research question \textbf{RQ2}: \textit{Can LLMs benefit the dataset construction in the \acs{MCPQA} task?} Delving into \textbf{RQ2}, on \OurData, we create a subset by randomly selecting some questions from each marketplace and perform GPT-4 auto-labeling.  %Specifically, we focus on two subtasks named \textbf{\ac{AG}} and \textbf{\ac{QR}}. 
Specifically, we focus on two widely-studied PQA subtasks under the multilingual cross-market settings, including  \textbf{\ac{AG}} \cite{wsdm19-pqa,wsdm19-pqa-chen} and \textbf{\ac{QR}} \cite{naacl21-pqa}.
For \acs{AG}, we ask LLMs to judge whether a question is answerable from associated reviews and provide its corresponding answer. We denote the subset as \OurDataans. For \acs{QR}, given two QA pairs, we ask LLMs to judge if one helps answer the other and denote the subset as \OurDatarank. With these two subsets, we conduct human assessment to analyze LLM-generated results from multiple perspectives. Surprisingly, in \OurDataans, 61.8\% LLM-generated answers are assumed `better' than the human ground truth.  
%  and to what extent  according to human annotators, LLM answers reached almost perfect scores in terms of accuracy, fluency, completeness, relavance.
% \yifei{add some findings}
% As a result, the LLM-based set \OurDatasub contains ? questions in total, with the average of ? questions from 9 main marketplaces.

Finally, we are interested in answering the research question \textbf{RQ3}: \textit{In the multilingual context, how can we effectively leverage the unique features of cross-market information to enhance product-related 
question answering?} To this end, we perform experiments of models on \acs{AG} and \acs{QR} subtasks. For each task, we report the performance of state-of-the-art methods under single- and cross-market scenarios on both \OurData and the corresponding LLM-labeled subset. We benchmark methods ranging from traditional lexical models (\ie BM25) to 
%state-of-the-art LLMs \dy{they are not SOTA}
LLM-based approaches (\ie LLaMA-2, Flan-T5). We demonstrate the superiority of cross-market methods against their single-market counterparts on both subtasks. 
% We then perform comprehensive experiments to compare our model with state-of-the-art methods on two datasets to see the superiority of our models. \yifei{add some findings and analyses} and to what extent 

In conclusion, our contributions are as follows:
\begin{itemize}
    \item We propose a novel task named MCPQA, where product-related information from an auxiliary marketplace is leveraged to answer questions in a resource-scarce marketplace in a multilingual setting. Specifically, we investigate two subtasks named \acs{AG} and \acs{QR}.
    \item We benchmark a large-scale real-world dataset to facilitate the research in the MCPQA task. We also collect two LLM-annotated subsets and adopt human assessment to analyze their characteristics.
    % \item To investigate the performance of LLM of the MCPQA task, we collect a GPT-4 based subset of our dataset and adopt human assessment to analyze its characteristics. We also report the performance of other LLMs on our task.
    \item To provide a comprehensive evaluation of the task and verify the superiority of cross-market methods, experiments are performed under both single/cross-market scenarios.\footnote{Data and code available in \url{https://github.com/yfyuan01/MCPQA}. }
\end{itemize}

\section{Related Work}
\header{Product-related QA}
% \subsection{Product-related QA} \citet{Yu2018ReviewAwareAP} adopt an aspect analytics model that learns aspects-related information to generate aspect-specific representations for new questions.
Product-related QA (PQA) seeks to address consumers' general inquiries by utilizing diverse product-related resources such as customer reviews, or the pre-existing QA sections available on a retail platform~\cite{emnlp12-pqa,Deng2023ProductQA}. Among the existing literature in this area, retrieval-based methods have been a popular direction that retrieve related reviews for providing the right answer~\cite{Wan2016ModelingAS,Zhang2019DiscoveringRR,Yu2018ReviewAwareAP,pkdd20-pqa,Zhang2020AnswerRF,emnlp20-pqa,www20-pqa}. For example, \citet{McAuley2015AddressingCA} propose a model that leverages questions from previous records for selecting the relevant review for the question. While most of these works assume there are no user-written answers available, \citet{Zhang2020AnswerRF} rank answers for the given question with review as an auxiliary input. Another line of research~\cite{wsdm19-pqa,wsdm19-pqa-chen,tois21-pqa,sigir21-pqa,cikm20-pqa,tois22-pqa} investigates answer generation grounding on retrieved product-related documents. More recently, \citet{Ghasemi2023CrossMarketPQ} introduce a novel task of utilizing available data in a resource-rich marketplace to answer new questions in a resource-scarce marketplace. Building upon their research, we expand the scope to a multilingual scenario, exploring additional marketplaces with non-English content. 
Furthermore, we explore both questions and review information from the auxiliary marketplace.

\header{Cross-domain and cross-lingual QA}
% \subsection{Cross-domain and cross-lingual QA}
Our work can be seen as a special format of cross-domain QA in E-commerce, which involves addressing questions that span different domains or fields of knowledge~\cite{coling18-crossdomain,Qu2020RocketQAAO,Liu2019XQAAC,Longpre2020MKQAAL,Yuan2021ConversationalFI,Abbasiantaeb2023LetTL}. For instance, \citet{Yu2017ModellingDR} propose a general framework that effectively applies the shared knowledge from a domain with abundant resources to a domain with limited resources. Also, cross-domain QA is often in close connection to cross-lingual QA in the sense that both involve transferring knowledge and understanding from one domain or language to another.~\cite{Artetxe2019OnTC,Clark2020TyDiQA,Zhang2019ImprovingLC}. \citet{Asai2020XORQC} expand the scope of open-retrieval question answering to a cross-lingual setting, allowing questions in one language to be answered using contents from another language. Most recently,~\citet{Shen2023xPQACP} introduce a multilingual PQA dataset called xPQA where cross-market information is also leveraged to aid the product-based question answering. 
% We provide a comprehensive comparison between \OurData and existing datasets in Appendix \ref{sec:appendix_comparison}. %Specifically, for a given question, the model first ranks related product page from an English marketplace, then generates some answers to present to the user based on the selected English candidate.

\section{Problem Formulation}
We investigate two subtasks of the MCPQA task, \textit{\acf{AG}} and \textit{\acf{QR}}, where answers to a product question are obtained by a generative or ranking way, respectively.
 
\header{Review-based answer generation} In this task, we assume that the answer can be obtained from the reviews of the product (or similar products). Based on the setting in ~\cite{Gupta2019AmazonQAAR}, we define this task in a multilingual cross-market scenario. Given a question $Q$ in the main marketplace $M_T$, we first retrieve and rank all the related reviews from similar items within both $M_T$ and auxiliary marketplace $M_A$. Given the retrieved review set $\Omega=\{R_1,...,R_k\}$, we predict if $Q$ is answerable from it by assigning a tag $t$. If yes, a generative function $\Gamma$ is learned: $A = \Gamma (Q,\Omega)$, so that answer $A$ is generated with both $Q$ and $\Omega$ as input.

\header{Product-related question ranking} Following the problem setting in \cite{Ghasemi2023CrossMarketPQ}, we assume that there are similar questions already asked about the product or similar products in other marketplaces. Therefore, given a \textit{main marketplace} in language $L_M$, denoted as $M_T$, which usually suffers from resource scarcity of the number of knowledgeable users answers, $M_T$ consists of several items $\{I_1, ..., I_m\}$, where each $I_k$ contains a set of question answering pairs $\{QA_{k1},...QA_{kn}\}$. Besides, there also exists a high-resource marketplace $M_A$, denoted as the \textit{auxiliary marketplace} (the \textbf{us} marketplace in our case) in language $L_A$ (note that in some cases $L_A$ can be the same as $L_M$). Similarly, $M_A$ also includes several items $\{I'_1, ..., I'_z\}$, where we can assume $z>>m$. The task is defined as, for a given question $Q$ in the main marketplace $M_T$, in a multilingual setting, we rank the questions from both $M_T$ and $M_A$ to take the corresponding answers of the top ranks as the possible answer to $Q$. 
\begin{table*}[!h]
\setlength{\abovecaptionskip}{5pt}   
\setlength{\belowcaptionskip}{0pt}
% \vspace{-1mm}
\small
    \centering
    \setlength{\tabcolsep}{1.5mm}
    \begin{tabular}{lrrrrrr}
    \toprule
        \textbf{Name} & \textbf{\# markets}  &  \textbf{\# languages} &   \textbf{\# products} & \textbf{\# questions} & \textbf{\# reviews} & \textbf{Average\ QPM}\\
        \midrule
    xPQA~\cite{Shen2023xPQACP} & 12 & 12 & 16,615  & 18,000 & - & 1,500\\
    XMarket-QA~\cite{Ghasemi2023CrossMarketPQ} & 2& 1 & 34,100&  4,821,332& - & 2,410,666 \\
    semiPQA~\cite{Shen2022semiPQAAS} & 1 & 1 & - & 11,243 & - & 11,243 \\
    SubjQA~\cite{Bjerva2020SubjQAAD} & 1 &1 & -  & 10,098 & 10,098 & 10,098 \\
    ReviewRC~\cite{Xu2019BERTPF} & 1 & 1 & - & 2,596 & 959 & 2,596 \\
    AmazonQA~\cite{Gupta2019AmazonQAAR} & 1 & 1 & 155,375 & 923,685 & 8,556,569 &923,685\\
    Amazon~\cite{McAuley2015AddressingCA} & 1 &  1 & 191,185 & 1,447,173  & 13,498,681 & 1,447,173\\
        \midrule
    Ours & 17 & 11 & 143,068 & 7,268,393 & 52,469,322 & 427,552\\
    
\bottomrule
    \end{tabular} 
    \caption{Comparison of our dataset with existing PQA datasets. QPM denotes question per marketplace.}
    \label{tab:comparison}
    \vspace{-3mm}
\end{table*}

\section{Data Collection \& Analysis}
We describe how we collect our dataset and perform several analysis to answer \textbf{RQ1} and \textbf{RQ2}.
\subsection{Data collection}
\subsubsection{Data preprocessing}
We construct our dataset on top of an Amazon product dataset called XMarket~\cite{Bonab2021CrossMarketPR}. XMarket includes authentic Amazon product metadata and user-generated reviews. Specifically, we sample 17 marketplaces covering 11 different languages from it. For each marketplace, we gather metadata and reviews for each product from XMarket. We also collect the question-answering pairs posed by the users by crawling the Amazon website. We then provide the corresponding English translation for the non-English contents of the Electronics category, naming it as \OurData. Specifically, we adopt the professional translation tool by DeepL\footnote{\url{https://www.deepl.com/}} for all the QA translation and the pre-trained NLLB model~\cite{team2022NoLL} fine-tuned on each non-English language for review translation. To the best of our knowledge, this is the first multilingual cross-market QA dataset with questions and reviews in the community (more information about privacy and license in Section \ref{ethics}).

\begin{figure}
\setlength{\abovecaptionskip}{5pt}   
\setlength{\belowcaptionskip}{0pt}
    \centering
    \includegraphics[width=\linewidth]{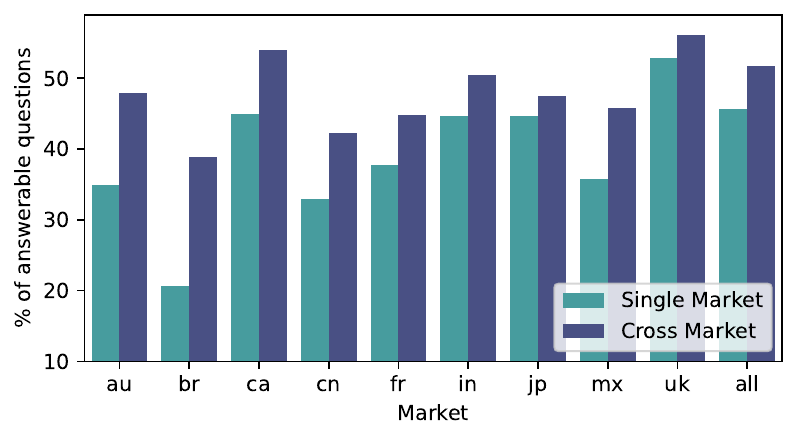}
    \caption{Portion of answerable questions in \OurData using single/cross-market review information.}
    \label{fig:anwerable_analysis}
    \vspace{-3mm}
\end{figure}
\subsubsection{LLM annotation}
For the two concerned subtasks, we both provide LLM-labeled data for supervised training. Specifically, we randomly select some data from  \OurData  and instruct GPT-4 to perform annotation. For \acs{AG}, we randomly select 1000 questions per marketplace.\footnote{For the \textbf{au} marketplace, the total is 584 questions, so we sample all of them.} Then, we follow the typical top-K pooling technique~\cite{Gonzlez2007TRECEA} and pool the top five retrieved reviews from a variety of retrieval methods. Next, we instruct GPT-4 to evaluate whether the question is answerable. If it is, GPT-4 generates an appropriate response using the question and reviews as input. If no, GPT-4 is instructed to output the reason and `no answer'. We denote this subset as \OurDataans. For QR, we randomly select 200 questions from each marketplace. Employing the same strategy, we retrieve the top five related question-answering pairs from both the main and auxiliary marketplaces. Consequently, we acquire 1,000 question-answering pairs for each marketplace, with 9k pairs in total. Then, GPT-4 is instructed to determine if the retrieved QA pairs would be useful in answering the original question by assigning a score from 0--2, representing `\textit{Very useful}', `\textit{Partially useful}', and `\textit{Not useful}', respectively.  We denote this subset as \OurDatarank. More details of the subsets as well as the prompts we gave to GPT-4 are listed in Appendix \ref{sec:appendix}.

% \subsubsection{Quality assurance}
\subsection{Data analysis}
% We first report the overall statistics of \OurData. Then to answer \textbf{RQ1}, we show that cross-market information has the potential to help get better answers from the dataset perspective. After that, we perform a temporal gap analysis of different marketplaces. Finally, to answer \textbf{RQ2}, some human evaluation is adopted to assess the quality of GPT-4 generated data.
\begin{figure}
\setlength{\abovecaptionskip}{5pt}   
\setlength{\belowcaptionskip}{0pt}
    \centering
    \includegraphics[width=0.75\linewidth]{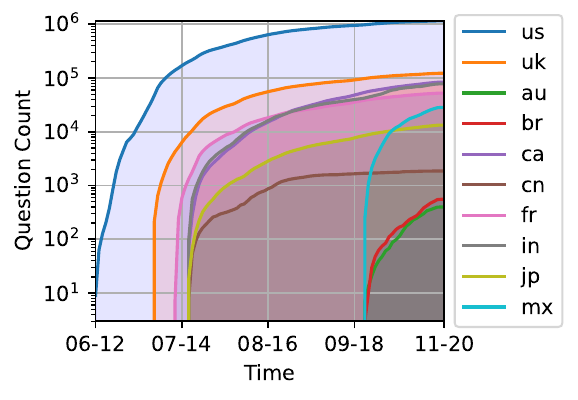}
    \caption{Temporal gap analysis. }
    \label{fig:time}
    \vspace{-3mm}
\end{figure}
\subsubsection{Dataset overview}
% Table~\ref{statistics} shows the overall statistics of \OurData. 
Overall, our dataset covers marketplaces ranging from those with a small scale (\ie \textbf{au}, \textbf{br}) to those with rich resources (\ie \textbf{uk}, \textbf{us}). It contains over 7 million product-related questions, 52 million reviews, and 143k unique products in total.

We compare our dataset with existing PQA datasets. According to Table \ref{tab:comparison},  our dataset exhibits advantages in various aspects: (1) \textbf{contains multiple languages} -- we provide product, question, and review information in the original text of their respective marketplaces and additionally offer the corresponding English translations; (2) \textbf{supports cross-market QA} -- our dataset is designed to facilitate question answering research across different marketplaces, enhancing its utility for cross-market analyses and evaluations; (3) \textbf{includes diverse information} -- compared with existing multilingual PQA dataset, \OurData encompasses comprehensive question and review information, paving the way for more diverse research avenues and tasks in the future; (4) \textbf{is large in scale} -- overall, \OurData surpasses most PQA datasets in terms of size, ensuring it comprises a substantial amount of data for experimentation and analysis. 
% We put the detailed statistics of the whole dataset as well as the \OurData labeled set in Appendix \ref{sec:appendix_statistics}.
\begin{table*}[t]
 %For the raw set, we include them in \OurData but leave the discussion to future work in Appendix \ref{sec:appendix_future}. \textbf{raw}
 \begin{adjustbox}{max width=0.99\textwidth}
    \setlength{\tabcolsep}{1mm}{
\begin{tabular}{lllllllllllll}
\hline
                        & \textbf{au} & \textbf{br} & \textbf{ca} & \textbf{cn}  & \textbf{fr} & \textbf{in} & \textbf{jp} & \textbf{mx} & \textbf{uk} & \textbf{us}  & Total\\
\hline
Language                &  en  & pt   &  en  &  cn  &  fr  & en   &  jp  &  es  &  en  & en   & - \\
\hline
% \textit{\OurData} &    &    &    &    &    &    &    &    &    &    \\
% \hline
Question Num.               & 584   &  1,378  & 101,126   & 3,324 &      66,536 &  115,829  & 17,418   &  34,433  &  164,848  &  1,782,092 &  2,287,568  \\
Review Num.               & 3,062   & 3,650   &  575,052  &  1,893   & 359,703   &  240,167  &  130,604  &  125,317  &  775,900  &   4,169,476&   6,384,824\\
Product Num.              &  85  & 95   &  5,432  & 210    & 2,199   & 2,085   &  903  &  1,464  & 4,406   &  29,976 &  30,606  \\
% Median ques. per item \\
Mean ques. len & 12.0$\pm$6.6 & 10.3$\pm$6.4 & 12.7$\pm$7.3& 10.2$\pm$8.1   & 15.2$\pm$6.9 & 10.1$\pm$6.0 & 20.3$\pm$15.7 & 10.9$\pm$6.8 & 13.6$\pm$7.6 & 13.4$\pm$7.8  & 13.3$\pm$7.9 \\
Medium ques. len. & 10 & 8 & 10 & 8  & 14 & 8 & 14 & 9 & 11 & 11  &   11 \\
Mean review len. & 25.5$\pm$30.0 & 17.5$\pm$25.3 & 29.9$\pm$50.4 & 56.4$\pm$60.2  & 39.1$\pm$49.7 &  21.8$\pm$42.9 & 28.7$\pm$36.8 & 28.7$\pm$36.8 & 40.1$\pm$68.4 & 59.3$\pm$93.0 & 51.5$\pm$84.0 \\
Medium review len. & 15 & 10 & 14 & 39  & 26 & 10 & 46 & 21 & 20 & 30 & 26 \\
% US Product Share &\\
% \OurDataans                       &    &    &    &    &    &    &    &    &    &   \\
% Question Num. & 584 & 1000 & 1000 & 1000 & 1000 & 1000 & 1000 & 1000 & 1000 & 1000 \\
% Review Num. \\
% Product Num.\\
\hline
\end{tabular}}
\end{adjustbox}
\caption{Overall statistics of the \OurData dataset. The length is reported on the token level.}
\label{statistics}
\end{table*}

\subsubsection{Dataset Statistics}
\label{sec:appendix_statistics}
Our full dataset contains product information from 17 different marketplaces, \textbf{au}, \textbf{br}, \textbf{ca}, \textbf{cn}, \textbf{de}, \textbf{es}, \textbf{fr}, \textbf{it}, \textbf{in}, \textbf{jp}, \textbf{mx}, \textbf{nl}, \textbf{sa}, \textbf{sg}, \textbf{tr}, \textbf{uk}, \textbf{us} respectively, covering 11 languages including \textbf{en}, \textbf{ar}, \textbf{cn}, \textbf{de}, \textbf{es}, \textbf{fr}, \textbf{it}, \textbf{nl}, \textbf{jp}, \textbf{pt}, \textbf{tr}. To reduce costs and facilitate baseline model training, we automatically translate the non-English contents in the Electronics category and abandon marketplaces with insufficient QA pairs. We name it as \OurData. Table \ref{statistics} shows the detailed statistics of \OurData.

\subsubsection{Cross-market QA analysis}
To answer \textbf{RQ1}, we compare the effect of product-related resources (\ie reviews) on question answering under both single- and cross-market scenarios. Figure~\ref{fig:anwerable_analysis} shows the comparison of answerable questions based on both single- and cross-market retrieved reviews in \OurData.\footnote{We adopt the answerable question prediction model in ~\cite{Gupta2019AmazonQAAR} to predict if a question is answerable or not given the review information.} We notice that the portion of answerable questions gets raised in every marketplace with cross-market reviews, with a particularly significant uplift observed in low-resource marketplaces (\ie \textbf{br}). This verifies the transferability of knowledge across marketplaces and underscores the advantages of leveraging cross-market information in enhancing the performance of product QA models. 

% In product-related question ranking, we compare the answers by ranking questions from the mainmarketplace(single-market) or the mixture of main/auxiliarymarketplace(cross-market). In review-based answer generation, we also compare the answers by taking the reviews from the main marketplace only (single-market) or the mixture of main/auxiliary marketplace (cross-market).
% \yifei{Human judgment, portion of answerable question, portion of relevant question} In reality, an emerging and resource-poor marketplace is supposed to benefit from a mature and resource-rich marketplace within the same period. 

\begin{table}[t]
\setlength{\abovecaptionskip}{5pt}   
\setlength{\belowcaptionskip}{0pt}
 \begin{adjustbox}{max width=0.48\textwidth}
    \setlength{\tabcolsep}{1mm}{
\begin{tabular}{ccccc}
\hline
& Very Bad & Bad &  Good & Very Good \\
\hline
Correctness  & 2.5  &   0.9    &  8.5  &  88.1\\
Completeness &   4.9  &  1.3        &    15.6  &  78.2 \\
Relevance &  3.5   &  2.7            & 13.4   & 80.4\\
Naturalness & 0.8 & 0.9 &  5.4 & 92.9 \\
\hline
\multicolumn{4}{l}{\textbf{Better than Ground Truth}} &  \textbf{61.8}\\
\hline
\end{tabular}}
\end{adjustbox}
\caption{Human evaluation on \OurDataans. All the numbers are shown in percentage. }
\label{human1}
\vspace{-3mm}
\end{table}

% \subsubsection{Temporal gap analysis}
We further analyze the temporal characteristics of \OurData.  Figure~\ref{fig:time} illustrates the cumulative sum of the number of QA data available on all the items in all marketplaces. There are several notable observations: 1) at the beginning, all marketplaces feature very few QA data. 2) At each timestep, the most resource-rich marketplace (\ie \textbf{us}) always dominates the number of QA data compared to other marketplaces by several orders of magnitude. 3) Over time, the resource intensity levels of different marketplaces continue to change. For example, the number of QA data in \textbf{mx} surpasses that in \textbf{cn} and \textbf{jp} after 2018/09. We further observe that, on average, over 70\% of the questions in the main marketplace have already been answered in the \textbf{us} auxiliary marketplace under the same item, before the first question even receives an answer. These findings confirm the practicality and importance of exploring how auxiliary marketplaces can be utilized as valuable resources for PQA. 
%\yifei{@Yang should be in charge of this part}

\subsubsection{LLM-generated data analysis}
To assess the quality of LLM-generated data, we perform several analyses. On both \OurDataans and \OurDatarank, we randomly select 50 questions from each marketplace, and hire 3 crowd-workers\footnote{We hire the crowd-workers via a professional data management company named Appen (https://appen.com/).} to manually assess the GPT-4 labels.
\begin{table}[t]
\setlength{\abovecaptionskip}{5pt}   
\setlength{\belowcaptionskip}{0pt}
 \begin{adjustbox}{max width=0.48\textwidth}
    \setlength{\tabcolsep}{1mm}{
\begin{tabular}{cccc}
\hline
& Incorrect & Partially correct &  Correct \\
\hline
Portion  & 6.0  &   10.9    &  83.0   \\

\hline
\multicolumn{3}{l}{Overall Precision} &  98.2\\
\multicolumn{3}{l}{Overall Recall} &  97.4\\
\multicolumn{3}{l}{\textbf{Overall F1}} &  \textbf{97.6}\\
\hline
\end{tabular}}
\end{adjustbox}
\caption{Human evaluation on \OurDatarank. All the numbers are shown in percentage. }
\label{human2}
\vspace{-3mm}
\end{table}

% Please add the following required packages to your document preamble:
% \usepackage{multirow}
\begin{table*}[h]
 \begin{adjustbox}{max width=0.99\textwidth}
    \setlength{\tabcolsep}{1mm}{
\begin{tabular}{clccccccccccccccccccccccccccccc}
\hline
                   & \multicolumn{1}{c}{\multirow{2}{*}{Method}} & \multicolumn{2}{c}{\textbf{au}}                        & \multicolumn{1}{c}{} & \multicolumn{2}{c}{\textbf{br}}                        & \multicolumn{1}{c}{} & \multicolumn{2}{c}{\textbf{ca}}                        & \multicolumn{1}{c}{} & \multicolumn{2}{c}{\textbf{cn}}                        & \multicolumn{1}{c}{} & \multicolumn{2}{c}{\textbf{fr}}                        & \multicolumn{1}{c}{} & \multicolumn{2}{c}{\textbf{in}}                        & \multicolumn{1}{c}{} & \multicolumn{2}{c}{\textbf{jp}}                        & \multicolumn{1}{c}{} & \multicolumn{2}{c}{\textbf{mx}}                        & \multicolumn{1}{c}{} & \multicolumn{2}{c}{\textbf{uk}}                        &
                   \multicolumn{1}{c}{} & \multicolumn{2}{c}{AVG} \\ \cline{3-4} \cline{6-7} \cline{9-10} \cline{12-13} \cline{15-16} \cline{18-19} \cline{21-22} \cline{24-25} \cline{27-28} \cline{30-31}
                   & \multicolumn{1}{c}{}                        & \multicolumn{1}{c}{B} & \multicolumn{1}{c}{R} & \multicolumn{1}{c}{} & \multicolumn{1}{c}{B} & \multicolumn{1}{c}{R} & \multicolumn{1}{c}{} & \multicolumn{1}{c}{B} & \multicolumn{1}{c}{R} & \multicolumn{1}{c}{} & \multicolumn{1}{c}{B} & \multicolumn{1}{c}{R} & \multicolumn{1}{c}{} & \multicolumn{1}{c}{B} & \multicolumn{1}{c}{R} & \multicolumn{1}{c}{} & \multicolumn{1}{c}{B} & \multicolumn{1}{c}{R} & \multicolumn{1}{c}{} & \multicolumn{1}{c}{B} & \multicolumn{1}{c}{R} & \multicolumn{1}{c}{} & \multicolumn{1}{c}{B} & \multicolumn{1}{c}{R} & \multicolumn{1}{c}{} & \multicolumn{1}{c}{B} & \multicolumn{1}{c}{R} &
                   \multicolumn{1}{c}{} & \multicolumn{1}{c}{B} & \multicolumn{1}{c}{R} \\ \hline
\multirow{4}{*}{\rotatebox{90}{Single}} &   BM25                                         &        6.1               &               7.0        &                      &    4.9                   &        6.9               &                      &                6.9       &           7.7           &                      &                4.8       & 5.2                      &                      &      8.0                 &    8.1                   &                      &         4.7              &                5.9       &                      &               11.0     &     9.6                  &                      &     7.0                  &      8.2                 &                      &    10.3                      &   9.3   & & 8.0  & 7.9                \\
&   BERT                                          & 7.4                 & 7.3                 &                      &   9.0               & 5.3                 &                      & 7.3                 & 6.8                 &                      &      5.4            &    5.0              &                      & 8.5                & 7.2                 &                      & 5.1                 & 4.8                 &                      & 10.6                 &      8.7 &                      & 9.4                  & 7.7                 &                      & 9.5                 & 8.2     &  & 7.9 & 7.0            \\ 
% &   mBERT                                          & 9.3                & 7.2                 &                      &              7.8    &      4.7            &                      & 7.6                 & 6.7                 &                      &          13.1        & 6.3                &                      & 7.7                 & 6.9                 &                      & 5.5                 & 4.8                 &                      &      15.9         &       10.1         &                      &      8.6            &     7.2            &                      & 9.8                 & 8.1   & & 8.2 &      6.9         \\ 
                   & T5                                  &         15.5        &        11.4              &                      & 14.3                 &            12.6           &                      & 16.4                &   12.1                    &                      &     13.5                 &          10.7             &                      &    16.5                   &               11.5        &                      &         12.8              &              9.9         &                      &                  22.6       &    15.6                   &                      &      20.2                 &                14.4       &                      &             18.9          &  13.3  &  & 16.9 & 12.2                   \\  
                   % &  mT5                               &    6.2                   &       5.3                &                      &            8.1           &         9.2              &                      &             14.3          &    10.0                   &                      &     19.5                  &             11.8        &                     &           15.5             &            10.7           &                      &                 9.7      &      8.7                 &                      &    26.3                   &           13.3            &                      &                   12.2     &        9.4              &                      &              14.6         &         9.6   & & 13.7 & 9.7          \\ 
                  
        &  Llama-2*                              &            \tinyincreasebg{10.2}            &    \tinyincreasebg{14.7}                   &        \tinyincreasebg{}              &     \tinyincreasebg{16.4}
                    &            \tinyincreasebg{17.1}           & \tinyincreasebg{}                     &                 \tinyincreasebg{15.9}      &    \tinyincreasebg{13.1}                   &     \tinyincreasebg{}                 &     \tinyincreasebg{14.8}                  &      \tinyincreasebg{13.6}                 &     \tinyincreasebg{}                 &    \tinyincreasebg{18.3}                   &            \tinyincreasebg{14.2}           &           \tinyincreasebg{}           &           \tinyincreasebg{13.5}          &      \tinyincreasebg{13.1}                   &       \tinyincreasebg{}               &               \tinyincreasebg{26.6}      &       \tinyincreasebg{19.7}                  &            \tinyincreasebg{}          &                \tinyincreasebg{22.3}       &         \tinyincreasebg{16.6}                &         \tinyincreasebg{}           &         \tinyincreasebg{20.1}              &  \tinyincreasebg{18.3} & \tinyincreasebg{}  &  \tinyincreasebg{17.8} & \tinyincreasebg{15.4}\\ 
                     % &  Mixtral                              &                       &                       &                      &                       &                       &                      &                       &                       &                      &                       &                       &                      &                       &                       &                      &                       &                       &                      &                       &                       &                      &                       &                       &                      &                       &  \\ 
                     \hline

\multirow{6}{*}{\rotatebox{90}{Cross}} &   BM25                                          & 10.6                & 7.9                 &                     &    9.0           &   6.1              &                      & 7.8                  & 7.9                  &                      & 4.6                 & 5.4                 &                      & 9.0                & 8.2                 &                      &     5.6             &      6.1            &                      & 11.3                 & 9.5                 &                      & 9.9                 & 9.1                &                      & 10.4                 & 9.2    & &  8.9 & 8.0             \\
      &   BERT                                          &     10.5                  &   8.1                    &                      &            9.5           &       6.4                &                    &     8.5                    &              8.9         &                      &           5.8            &            5.1           &                      &              9.8         &    8.3                   &                     &          6.1              &         7.3              &                      &    11.8                   &         9.6              &                      &               10.4        &    8.7                   &                      &      11.4                 &  10.3       & & 9.4 &  9.0             \\

               &     Exact-T5        &   14.0                  
     &        11.8               &                      &         16.6              &     13.0                  &                      &         18.2              &               11.9        &                      &      13.0                 &             11.0          &                     &                 18.1       &        11.3               &                      &            12.5           &         10.1              &                      &            22.7           &      15.0                 &                      &              20.3         &   14.2                    &                    &    20.6                   &   13.7   &  & 17.9 & 12.3  \\                     
        &     T5               
        &           \smallincreasebg{16.1}            &    \smallincreasebg{11.3}                  &       \smallincreasebg{}               &   \smallincreasebg{17.0}                    &            \smallincreasebg{14.1}           &       \smallincreasebg{}               &       \smallincreasebg{17.0}                &            \smallincreasebg{12.7}           &               \smallincreasebg{}       &       \smallincreasebg{15.1}                &      \smallincreasebg{11.3}                 &      \smallincreasebg{}                &   \smallincreasebg{19.4}                    &             \smallincreasebg{12.6}          &   \smallincreasebg{}                     &       \smallincreasebg{13.2}                &         \smallincreasebg{10.6}              &      \smallincreasebg{}                &   \smallincreasebg{23.6}                  &           \smallincreasebg{16.0}              &   \smallincreasebg{}                   &   \smallincreasebg{22.3}                    &       \smallincreasebg{16.6}                &    \smallincreasebg{}                  &       \smallincreasebg{20.2}                &  \smallincreasebg{15.4} &\smallincreasebg{} & \smallincreasebg{18.1} & \smallincreasebg{13.5}    \\
         &  Exact-Llama-2*                             &                19.5      &     15.1                 &                     &     17.4                   &         15.5              &                      &            16.4           &          13.8             &                      &    15.6                 &                11.4         &                      &            21.6           &  17.6                     &                      &  \textbf{16.9}                     &              \textbf{15.1}         &                      &             27.3          &              17.8           &                    &                24.7       &     17.8                  &                      &               22.4        &   19.8  & & 20.1 & 17.0  \\  
% \rowcolor{mygray}
 &     Llama-2*  &  \smallincreasebg{\textbf{21.4}}  &  \smallincreasebg{\textbf{20.6}}  & \smallincreasebg{} & \smallincreasebg{\textbf{18.9}} & \smallincreasebg{\textbf{19.5}} & \smallincreasebg{} &  \smallincreasebg{\textbf{19.5}} & \smallincreasebg{\textbf{14.4}}  & \smallincreasebg{} &  \smallincreasebg{\textbf{17.6}}  & \smallincreasebg{\textbf{15.5}}  & \smallincreasebg{} &  \smallincreasebg{\textbf{22.0}}  &  \smallincreasebg{\textbf{19.0}}  & \smallincreasebg{}   & \smallincreasebg{16.5}  & \smallincreasebg{15.0}   &  \smallincreasebg{} & \smallincreasebg{\textbf{29.5}} & \smallincreasebg{\textbf{18.6}} & \smallincreasebg{}  & \smallincreasebg{\textbf{25.7}} &  \smallincreasebg{\textbf{19.2}}  & \smallincreasebg{} &  \smallincreasebg{\textbf{25.0}}  &  \smallincreasebg{\textbf{22.7}} & \smallincreasebg{} & \smallincreasebg{\textbf{21.7}} &  \smallincreasebg{\textbf{18.3}}
                  \\
                   \hline
\end{tabular}}
\end{adjustbox}
\caption{Experimental results of \acs{AG} on \OurData. Where B denotes BLEU-4, R denoted ROUGE-L. * denotes LLM based methods. The best-performed model in the single-market setting is highlighted in light grey. The models in dark grey are highlighted to distinguish from their Exact- counterparts. }
\label{exp1}
\end{table*}
% Please add the following required packages to your document preamble:
% \usepackage{multirow}
\begin{table*}[]
 \begin{adjustbox}{max width=0.99\textwidth}
    \setlength{\tabcolsep}{1mm}{
\begin{tabular}{llccccccccccccccccccccccccccccc}
\hline
                   & \multicolumn{1}{c}{\multirow{2}{*}{Method}} & \multicolumn{2}{c}{\textbf{au}}                        & \multicolumn{1}{c}{} & \multicolumn{2}{c}{\textbf{br}}                        & \multicolumn{1}{c}{} & \multicolumn{2}{c}{\textbf{ca}}                        & \multicolumn{1}{c}{} & \multicolumn{2}{c}{\textbf{cn}}                        & \multicolumn{1}{c}{} & \multicolumn{2}{c}{\textbf{fr}}                        & \multicolumn{1}{c}{} & \multicolumn{2}{c}{\textbf{in}}                        & \multicolumn{1}{c}{} & \multicolumn{2}{c}{\textbf{jp}}                        & \multicolumn{1}{c}{} & \multicolumn{2}{c}{\textbf{mx}}                        & \multicolumn{1}{c}{} & \multicolumn{2}{c}{\textbf{uk}}                        &
                   \multicolumn{1}{c}{} & \multicolumn{2}{c}{AVG} \\ \cline{3-4} \cline{6-7} \cline{9-10} \cline{12-13} \cline{15-16} \cline{18-19} \cline{21-22} \cline{24-25} \cline{27-28} \cline{30-31}
                   & \multicolumn{1}{c}{}                        & \multicolumn{1}{c}{B} & \multicolumn{1}{c}{R} & \multicolumn{1}{c}{} & \multicolumn{1}{c}{B} & \multicolumn{1}{c}{R} & \multicolumn{1}{c}{} & \multicolumn{1}{c}{B} & \multicolumn{1}{c}{R} & \multicolumn{1}{c}{} & \multicolumn{1}{c}{B} & \multicolumn{1}{c}{R} & \multicolumn{1}{c}{} & \multicolumn{1}{c}{B} & \multicolumn{1}{c}{R} & \multicolumn{1}{c}{} & \multicolumn{1}{c}{B} & \multicolumn{1}{c}{R} & \multicolumn{1}{c}{} & \multicolumn{1}{c}{B} & \multicolumn{1}{c}{R} & \multicolumn{1}{c}{} & \multicolumn{1}{c}{B} & \multicolumn{1}{c}{R} & \multicolumn{1}{c}{} & \multicolumn{1}{c}{B} & \multicolumn{1}{c}{R} &
                   \multicolumn{1}{c}{} & \multicolumn{1}{c}{B} & \multicolumn{1}{c}{R} \\ \hline
\multirow{4}{*}{\rotatebox{90}{Single}} &   BM25   & 10.3 & 11.7 & &  10.7   & 12.5   &   &  8.3   & 13.0 & &   8.5  &  10.1   & & 11.6   &  15.7  & &   11.7 &14.3 & &   12.8   &  12.1     & &  13.3   & 13.6     & &   12.4 & 14.7   & &     10.7      & 13.3                                        \\
&   BERT &    12.4 & 10.0 & & 14.8 & 8.7 & & 11.3 & 8.8 & & 8.5 & 7.1 & &  11.1 & 10.2 & & 12.0 & 10.6 & & 10.9 & 9.0 & & 14.1 & 9.5 &  & 9.0 & 11.1 & &    10.8 & 9.5 \\ 
% &   mBERT    & 13.4 & 9.5 & & 14.0 & 8.9 & & 10.9 & 8.9 && 15.7 & 9.9 && 12.1 & 9.6 && 13.0 & 10.7 && 14.5 & 12.8  && 17.0 & 11.0 & & 8.5 & 11.5 & & 13.2 & 10.4\\ 
& T5     &   29.8                   &       27.0                &       &  26.7      & 33.6      &        &   29.2   &  27.4      &       &     31.1   &    24.2    &        &  34.9     &   30.8  &       &  29.0    &   32.2    &      &   31.1    &   27.0   &       &   27.2   &  26.5     &      &  29.5     &  25.9  & & 29.9 & 28.4           \\  
% &  mT5    & 10.6 & 14.3 &  & 5.2 & 13.5  & & 6.8 & 10.4& & 41.1 & 26.4 & & 19.9 & 17.4 & & 9.2 & 14.7 &   & 34.2 &29.1 & & 24.5  & 16.3 & & 7.2 & 13.5 & & 18.0 & 17.4    \\ 
&     Llama-2*                                        &  \tinyincreasebg{35.7}  &  \tinyincreasebg{34.3}  & \tinyincreasebg{}  & \tinyincreasebg{37.6}  &  \tinyincreasebg{\textbf{40.8}} & \tinyincreasebg{}  &  \tinyincreasebg{36.3} & \tinyincreasebg{37.2} & \tinyincreasebg{} & \tinyincreasebg{38.7}  & \tinyincreasebg{34.3} &\tinyincreasebg{}  & \tinyincreasebg{35.7} & \tinyincreasebg{32.6} & \tinyincreasebg{} & \tinyincreasebg{34.4}  &\tinyincreasebg{35.8} & \tinyincreasebg{}  &  \tinyincreasebg{34.7} & \tinyincreasebg{32.4}  & \tinyincreasebg{} & \tinyincreasebg{35.9} & \tinyincreasebg{34.7} & \tinyincreasebg{} & \tinyincreasebg{35.4}      &  \tinyincreasebg{37.0} & 
 \tinyincreasebg{} & \tinyincreasebg{35.4}  & \tinyincreasebg{35.9}   \\            

                     % &  Mixtral                              &                       &                       &                      &                       &                       &                      &                       &                       &                      &                       &                       &                      &                       &                       &                      &                       &                       &                      &                       &                       &                      &                       &                       &                      &                       &  \\ 
                     \hline

\multirow{6}{*}{\rotatebox{90}{Cross}} &    BM25  &  13.5 & 11.0    & &  12.9   &  10.0   & & 13.4 &  12.2 & &  7.4  &  8.5   & &  12.8  & 13.0   & & 14.6  & 15.0 & &  11.6    &  10.1    & &       15.5  &   12.6    & & 12.0 & 15.2 & &   12.6  & 12.0\\
                   &   BERT  & 15.8 & 10.6 & & 15.7 & 11.0  & & 14.4 & 9.8 & & 6.8& 8.1 & & 12.2 & 14.2 & & 13.0 & 12.1 &  &13.8 & 11.3 & &   15.7 & 11.1 & & 10.1 & 13.1 & & 12.9 &   11.3                                                \\
   &     Exact-T5    & 30.9  & 28.2& &30.1 & 29.0& & 29.3 & 30.7 & & 29.8 &  26.7 && 34.7 & 31.7  & & 31.8 & 30.3 & & 30.0 &24.6   & & 27.3 & 28.0 & &29.1 & 25.9  & & 30.3 & 28.4                        \\   
                  
     &     T5     &   \smallincreasebg{32.0}  &\smallincreasebg{30.2}            &                \smallincreasebg{}      &              \smallincreasebg{31.0}         &         \smallincreasebg{28.6}              &           \smallincreasebg{}           &           \smallincreasebg{29.9}            &      \smallincreasebg{29.7}                 &             \smallincreasebg{}         &     \smallincreasebg{32.1}                  &         \smallincreasebg{26.8}             &                  \smallincreasebg{}    &               \smallincreasebg{32.2}        &        \smallincreasebg{31.5}               &    \smallincreasebg{}                  &                \smallincreasebg{30.1}        &         \smallincreasebg{32.4}             &      \smallincreasebg{}              &    \smallincreasebg{36.3}                   &        \smallincreasebg{29.9}               &          \smallincreasebg{}            &        \smallincreasebg{29.4}               &             \smallincreasebg{27.6} &   \smallincreasebg{}                   &               \smallincreasebg{30.2}        &  \smallincreasebg{26.0}  & \smallincreasebg{} & \smallincreasebg{31.4} & \smallincreasebg{29.1} 
                   \\
  &  Exact-Llama-2*   & \textbf{37.0}  & 34.6 &   &   34.1  & 32.6  & & 38.0   & 39.9 &  & 33.0  & 35.2 &  &  \textbf{40.8} & \textbf{44.3} & & 36.2 & 40.2 & & 38.0  &  34.7 &  &  38.4
                & 37.8 &   &  35.2 & 37.9 & & 36.7 & 37.3 \\  
                               
&  Llama-2* &  \smallincreasebg{35.9} & \smallincreasebg{\textbf{37.4}} & \smallincreasebg{}  &   \smallincreasebg{\textbf{38.0}}  &  \smallincreasebg{37.9} & \smallincreasebg{} &  \smallincreasebg{\textbf{39.2}}  & \smallincreasebg{\textbf{40.2}} & \smallincreasebg{}  &  \smallincreasebg{\textbf{39.1}} & \smallincreasebg{\textbf{36.9}} &  \smallincreasebg{} &  \smallincreasebg{39.6} & \smallincreasebg{41.7} & \smallincreasebg{} & \smallincreasebg{\textbf{37.0}} & \smallincreasebg{\textbf{41.0}} & \smallincreasebg{} & \smallincreasebg{\textbf{40.9}}  &\smallincreasebg{\textbf{35.2}} &\smallincreasebg{} & \smallincreasebg{\textbf{38.8}}   & \smallincreasebg{\textbf{37.1}} & \smallincreasebg{}  & \smallincreasebg{\textbf{35.9}}  & \smallincreasebg{\textbf{38.5}} & \smallincreasebg{}& \smallincreasebg{\textbf{38.4}} & \smallincreasebg{\textbf{38.5}}\\                  
 \hline           
\end{tabular}}
\end{adjustbox}
\caption{Experimental results of \acs{AG} on \OurDataans.}
\label{exp1_gpt4}
\vspace{-3mm}
\end{table*}

\header{Review-based generation} For \OurDataans, we ask the crowd-workers to assess GPT-4-generated answers in terms of correctness, completeness, relevance, and naturalness. The detailed definitions of them are listed in Appendix \ref{sec:human_metrics}. For each metric, we asked them to assign a score from $-2$ to $+2$ to assess the answer quality, with $-2$ representing `very bad' and $+2$ representing `very good.' We also asked them to choose the better answer between the GPT-4 generated response and the human-provided ground truth, without disclosing the true category. From Table~\ref{human1}, we note that GPT-4 answers demonstrate reasonable performance in terms of every metric. 
Surprisingly, our findings reveal that in most cases, human assessors perceive GPT-4 results to be better than human-generated ground truth. Notably, GPT-4's outcomes are derived solely from reviews, whereas human ground truth relies on both reviews and user experiences. 

\header{Question ranking} For \OurDatarank, we ask the workers to judge the GPT-4-generated question ranking quality, by assigning a score between 0--2 to each sample, where 0 denotes GPT-4 answers are not correct, 1 as partially correct, and 2 as completely correct. Furthermore, we instruct the annotators to provide their own judgment of the ranking score if they mark GPT-4 answers as 0 or 1. Table \ref{human2} shows that the quality of the generated question ranking results by GPT-4 is also deemed satisfactory, achieving over 93\% correctness in question ranking pairs and an overall F1 score of 97.6\%.

% \input{Tables/exp1_new}
% Please add the following required packages to your document preamble:
% \usepackage{multirow}
\begin{table*}[h]
 \begin{adjustbox}{max width=0.99\textwidth}
    \setlength{\tabcolsep}{1mm}{
\begin{tabular}{llccccccccccccccccccccccccccccc}
\hline
                   & \multicolumn{1}{c}{\multirow{2}{*}{Method}} & \multicolumn{2}{c}{\textbf{au}}                        & \multicolumn{1}{c}{} & \multicolumn{2}{c}{\textbf{br}}                        & \multicolumn{1}{c}{} & \multicolumn{2}{c}{\textbf{ca}}                        & \multicolumn{1}{c}{} & \multicolumn{2}{c}{\textbf{cn}}                        & \multicolumn{1}{c}{} & \multicolumn{2}{c}{\textbf{fr}}                        & \multicolumn{1}{c}{} & \multicolumn{2}{c}{\textbf{in}}                        & \multicolumn{1}{c}{} & \multicolumn{2}{c}{\textbf{jp}}                        & \multicolumn{1}{c}{} & \multicolumn{2}{c}{\textbf{mx}}                        & \multicolumn{1}{c}{} & \multicolumn{2}{c}{\textbf{uk}}                        &
                   \multicolumn{1}{c}{} & \multicolumn{2}{c}{AVG} \\ \cline{3-4} \cline{6-7} \cline{9-10} \cline{12-13} \cline{15-16} \cline{18-19} \cline{21-22} \cline{24-25} \cline{27-28} \cline{30-31}
                   & \multicolumn{1}{c}{}                        & \multicolumn{1}{c}{M} & \multicolumn{1}{c}{P} & \multicolumn{1}{c}{} & \multicolumn{1}{c}{M} & \multicolumn{1}{c}{P} & \multicolumn{1}{c}{} & \multicolumn{1}{c}{M} & \multicolumn{1}{c}{P} & \multicolumn{1}{c}{} & \multicolumn{1}{c}{M} & \multicolumn{1}{c}{P} & \multicolumn{1}{c}{} & \multicolumn{1}{c}{M} & \multicolumn{1}{c}{P} & \multicolumn{1}{c}{} & \multicolumn{1}{c}{M} & \multicolumn{1}{c}{P} & \multicolumn{1}{c}{} & \multicolumn{1}{c}{M} & \multicolumn{1}{c}{P} & \multicolumn{1}{c}{} & \multicolumn{1}{c}{M} & \multicolumn{1}{c}{P} & \multicolumn{1}{c}{} & \multicolumn{1}{c}{M} & \multicolumn{1}{c}{P} &
                   \multicolumn{1}{c}{} & \multicolumn{1}{c}{M} & \multicolumn{1}{c}{P} \\ \hline
\multirow{4}{*}{\rotatebox{90}{Single}} &   BM25 & 24.5 & 16.9 & &15.2 & 18.3 & & 31.5 & 28.7 & & 22.0 & 28.7 &  & 21.0  & 34.7 & &44.4 & 46.0 & & 23.8 & 31.5 & & 28.9 & 38.7 & & 38.4 & 40.2 &  &27.7 & 31.5                                           \\
 % &mBERT & 25.9  & 33.0 &  & 16.1 & 26.7 & &32.7  & 33.5 & & 18.5 & 30.0 & & 17.9 & 31.2 & &45.2 & 46.2 & & 24.1 & 32.5 & & 32.8 & 40.2 & &39.9  & 43.7 & & 28.1 & 35.2 \\
 &BERT & 26.9  &  43.0 & & 18.2 & 35.0 &  &  30.4  & 42.8 &  & 18.2 & 34.3 &   &  17.7   &  40.8 &  &  47.9 & 52.7  &  &  28.5 & 34.2 & & 30.0 & 47.0 &  & 40.0 &51.8 & & 28.6 & 42.4\\
 % & CMJim &  & \\
 & UPR-m & 30.4 & 46.0 & & 21.9 & 39.3 & & 31.9 & 48.0 & & 36.2 & 45.5 & & 36.3 & 43.7 & & 25.7 & 56.3 & & 34.7 & 43.3 & & 39.5 & 54.2 & & 32.5 & 52.7 & & 32.1 & 47.7\\
 & UPR-l* & \tinyincreasebg{38.9} & \tinyincreasebg{48.8} & \tinyincreasebg{} & \tinyincreasebg{27.8} & \tinyincreasebg{43.3} &\tinyincreasebg{} & \tinyincreasebg{36.5} & \tinyincreasebg{49.7} & \tinyincreasebg{} & \tinyincreasebg{38.1} & \tinyincreasebg{48.3} & \tinyincreasebg{} & \tinyincreasebg{42.5} & \tinyincreasebg{47.3} & \tinyincreasebg{} & \tinyincreasebg{35.2} & \tinyincreasebg{59.8} & \tinyincreasebg{} & \tinyincreasebg{43.3} & \tinyincreasebg{47.2} &\tinyincreasebg{} & \tinyincreasebg{49.0} & \tinyincreasebg{57.2} & \tinyincreasebg{} & \tinyincreasebg{38.9} & \tinyincreasebg{55.5} & \tinyincreasebg{} & \tinyincreasebg{38.9} & \tinyincreasebg{50.8}\\
                   \hline
\multirow{7}{*}{\rotatebox{90}{Cross}} & BM25  &        51.2               &               45.2        &                      &           47.4            &        40.0               &                      &                51.0      &           47.5           &                      &                50.2      & 46.8                      &                      &      50.8                 &    44.3 &                      &                58.0       &                57.5       &                      &                54.6    &     45.5                  &                      &     59.0                  &      54.3                 &                      &    50.8 & 57.5 & & 52.6 & 48.7             \\
& Exact-BERT & 50.7 & 38.8 &  & 49.1 & 41.8 & &48.8 & 47.0 & & 46.2&46.5 &  &  50.1 & 44.7 & & 59.0 & 57.3 & &54.8 &45.8 & &59.3& 55.7 & & 51.2 & 57.3 & & 52.1 & 48.3\\
& BERT  & \smallincreasebg{52.3} &\smallincreasebg{45.7} & \smallincreasebg{} & \smallincreasebg{49.7} &  \smallincreasebg{42.8} & \smallincreasebg{} &  \smallincreasebg{50.4} &  \smallincreasebg{48.8} & \smallincreasebg{} & \smallincreasebg{49.3}   & \smallincreasebg{44.2}   & \smallincreasebg{}   & \smallincreasebg{49.4} &  \smallincreasebg{43.5} & \smallincreasebg{} & \smallincreasebg{60.5}  &  \smallincreasebg{58.3} & \smallincreasebg{} & \smallincreasebg{55.9} & \smallincreasebg{46.0} & \smallincreasebg{} & \smallincreasebg{59.7} & \smallincreasebg{57.0} & \smallincreasebg{} & \smallincreasebg{52.5} & \smallincreasebg{59.3} & \smallincreasebg{} & \smallincreasebg{53.3} & \smallincreasebg{49.5}\\
& CMJim &  57.5 &   56.7 & & 52.4  & 49.3 & & 53.3 & 57.7 & & 54.0 & 50.5 &  & 56.9 & 54.3  &  &62.9 &  66.8 & & 58.4 & 53.2 & & 64.9 & 63.8  &  & 52.9 &62.7 & & 57.0 & 57.2\\
& UPR-m & 59.1 & 55.5 & & \textbf{57.8} & 56.0 & & 54.3 & 58.5 & & 52.8 & 52.1 & & 54.9 & 52.3 & & 64.1 & 64.3 & & 57.5 & 52.9 & & 62.8 & 63.7 & & 53.6 & 64.5 & & 57.4 & 57.8 \\
& Exact-UPR-l* & 59.3 & 56.0 & & 56.3   & 57.1  & &\textbf{59.7} &59.5 &  &54.4 &53.7 & &  55.4 & 54.0 &  & 65.6  & 68.8 &  & 58.5 & 53.3&   &  62.4 &   62.9 &  &  54.1  & 62.8 & & 58.4 & 58.7\\
& UPR-l*  &\smallincreasebg{\textbf{60.0}} & \smallincreasebg{\textbf{59.5}} & \smallincreasebg{} &  \smallincreasebg{57.7} & \smallincreasebg{\textbf{57.5}} & \smallincreasebg{} &\smallincreasebg{59.0} & \smallincreasebg{\textbf{63.2}}& \smallincreasebg{} &\smallincreasebg{\textbf{61.1}}  & \smallincreasebg{\textbf{54.8}} &\smallincreasebg{} & \smallincreasebg{\textbf{57.8}} & \smallincreasebg{\textbf{58.0}} & \smallincreasebg{} & \smallincreasebg{\textbf{67.2}} &\smallincreasebg{\textbf{70.5}} &\smallincreasebg{} & \smallincreasebg{\textbf{62.8}} & \smallincreasebg{\textbf{56.0}} & \smallincreasebg{} &\smallincreasebg{\textbf{67.2}} & \smallincreasebg{\textbf{66.2}} & \smallincreasebg{} &\smallincreasebg{\textbf{59.0}} &\smallincreasebg{\textbf{66.3}} & \smallincreasebg{} & \smallincreasebg{\textbf{60.5}} & \smallincreasebg{\textbf{60.9}}\\

\hline
\end{tabular}}
\end{adjustbox}
\caption{Unsupervised experimental results of the \acs{QR} on \OurData. Where M and P denote MRR and Precision@3, respectively. * denotes LLM-based methods. }
\label{exp2}
\end{table*}

% Please add the following required packages to your document preamble:
% \usepackage{multirow}
\begin{table*}[h]
 \begin{adjustbox}{max width=0.99\textwidth}
    \setlength{\tabcolsep}{1mm}{
\begin{tabular}{clccccccccccccccccccccccccccccc}
\hline
                   & \multicolumn{1}{c}{\multirow{2}{*}{Method}} & \multicolumn{2}{c}{\textbf{au}}                        & \multicolumn{1}{c}{} & \multicolumn{2}{c}{\textbf{br}}                        & \multicolumn{1}{c}{} & \multicolumn{2}{c}{\textbf{ca}}                        & \multicolumn{1}{c}{} & \multicolumn{2}{c}{\textbf{cn}}                        & \multicolumn{1}{c}{} & \multicolumn{2}{c}{\textbf{fr}}                        & \multicolumn{1}{c}{} & \multicolumn{2}{c}{\textbf{in}}                        & \multicolumn{1}{c}{} & \multicolumn{2}{c}{\textbf{jp}}                        & \multicolumn{1}{c}{} & \multicolumn{2}{c}{\textbf{mx}}                        & \multicolumn{1}{c}{} & \multicolumn{2}{c}{\textbf{uk}}                        &
                   \multicolumn{1}{c}{} & \multicolumn{2}{c}{AVG} \\ \cline{3-4} \cline{6-7} \cline{9-10} \cline{12-13} \cline{15-16} \cline{18-19} \cline{21-22} \cline{24-25} \cline{27-28} \cline{30-31}
                   & \multicolumn{1}{c}{}                        & \multicolumn{1}{c}{M} & \multicolumn{1}{c}{P} & \multicolumn{1}{c}{} & \multicolumn{1}{c}{M} & \multicolumn{1}{c}{P} & \multicolumn{1}{c}{} & \multicolumn{1}{c}{M} & \multicolumn{1}{c}{P} & \multicolumn{1}{c}{} & \multicolumn{1}{c}{M} & \multicolumn{1}{c}{P} & \multicolumn{1}{c}{} & \multicolumn{1}{c}{M} & \multicolumn{1}{c}{P} & \multicolumn{1}{c}{} & \multicolumn{1}{c}{M} & \multicolumn{1}{c}{P} & \multicolumn{1}{c}{} & \multicolumn{1}{c}{M} & \multicolumn{1}{c}{P} & \multicolumn{1}{c}{} & \multicolumn{1}{c}{M} & \multicolumn{1}{c}{P} & \multicolumn{1}{c}{} & \multicolumn{1}{c}{M} & \multicolumn{1}{c}{P} &
                   \multicolumn{1}{c}{} & \multicolumn{1}{c}{M} & \multicolumn{1}{c}{P} \\ \hline
% \multirow{5}{*}{\rotatebox{90}{Single-market}} &   BM25                                         &        6.1               &               7.0        &                      &    4.9                   &        6.9               &                      &                6.9       &           7.7           &                      &                4.8       & 5.2                      &                      &      8.0                 &    8.1                   &                      &         4.7              &                5.9       &                      &               11.0     &     9.6                  &                      &     7.0                  &      8.2                 &                      &    10.3                    \\
\multirow{4}{*}{\rotatebox{90}{Single}} &BERT-f & 32.7 & 44.4 & & 25.8 & 48.9 & & 30.0 & 42.2 & &31.7 & 35.6 & & 45.8 & 47.8 & & 46.2 & 64.4 & & 51.1 & 48.9 & & 46.4 & 58.9 & & 54.4 & 61.1 & & 40.5 & 50.2\\
 % &mBERT & 32.8  & 41.1 & & 21.9 & 40.0 & &  27.5 & 40.0 & & 29.4 &34.4 & & 41.9 & 45.6 & & 42.9 & 56.7 & &   48.6 & 41.1 & & 42.3 & 51.1 & &  52.9& 56.7 & & 37.8 &  45.2\\
 % & CMJim \\
& T5  &  29.4 & 42.2 &  & 23.3 & 41.1 & &   31.7 &   38.9 & & 31.3 & 30.9  & & 42.0 & 45.1 & & 43.8  &  58.4 & &  49.7  & 47.8 &   &   44.4 & 54.1  &   &   53.9  & 56.4 & & 38.8 & 46.1\\
 & monoT5 & 30.1 & 44.4 & &23.1 & 41.1 & & 31.3 &43.2 & & 31.4 &31.1  & & 43.2 & 46.7 &  & 49.4 & 63.3  & & 53.5 & 49.9 & & 47.8 & 54.4 &  & 53.4 & 58.9 &  & 40.4 & 48.1\\

& Flan-T5* & \tinyincreasebg{39.7} & \tinyincreasebg{51.1} & \tinyincreasebg{} & \tinyincreasebg{26.9} & \tinyincreasebg{50.0} & \tinyincreasebg{} &  \tinyincreasebg{34.0} & \tinyincreasebg{46.7} & \tinyincreasebg{} & \tinyincreasebg{38.3} & \tinyincreasebg{42.2} & \tinyincreasebg{} & \tinyincreasebg{52.2} & \tinyincreasebg{54.4} & \tinyincreasebg{} & \tinyincreasebg{51.4} & \tinyincreasebg{63.3} & \tinyincreasebg{} & \tinyincreasebg{54.8} & \tinyincreasebg{64.4} & \tinyincreasebg{} & \tinyincreasebg{49.3} &  \tinyincreasebg{60.0} & \tinyincreasebg{} &   \tinyincreasebg{55.8} & \tinyincreasebg{62.2} & \tinyincreasebg{} & \tinyincreasebg{44.7} & \tinyincreasebg{54.9}\\

                   \hline
% \multirow{6}{*}{\rotatebox{90}{Cross-market}} & BM25 \\
\multirow{6}{*}{\rotatebox{90}{Cross}} & Exact-BERT-f & 46.4 & 45.6 & & 40.0 & 51.1 & & 51.5 & 47.8 & &49.4 & 45.6 & & 52.3 &53.2 & & 49.3 & 66.0 & & 53.4 & 47.8 & & 48.9 &63.3 & &58.7 & 66.7 & & 50.0 & 54.1 \\
&BERT-f & \smallincreasebg{58.6} & \smallincreasebg{54.4} & \smallincreasebg{} & \smallincreasebg{52.3} & \smallincreasebg{54.4} & \smallincreasebg{} & \smallincreasebg{55.3} & \smallincreasebg{53.3} & \smallincreasebg{} & \smallincreasebg{56.2} & \smallincreasebg{46.7} & \smallincreasebg{} & \smallincreasebg{53.9} & \smallincreasebg{55.6} & \smallincreasebg{} & \smallincreasebg{65.8} & \smallincreasebg{70.0} & \smallincreasebg{} & \smallincreasebg{56.0} & \smallincreasebg{52.2} & \smallincreasebg{} & \smallincreasebg{63.2} & \smallincreasebg{71.1} & \smallincreasebg{} & \smallincreasebg{59.6} & \smallincreasebg{70.0} & \smallincreasebg{} & \smallincreasebg{57.9} & \smallincreasebg{58.6}\\
% & CMJim \\
% & UPR \\
& Exact-monoT5 & 52.6 & 48.9 & & 50.7 & 53.8 & &54.6 & 55.6 & &  54.4 & 44.9 & & 53.2 & 53.1   & & 63.1  & 71.0  & & 56.9 & 52.1  & &  62.8 &  67.8 & &   59.3 & 66.8 & & 56.4 & 57.1  \\
& monoT5 & \smallincreasebg{52.9} & \smallincreasebg{53.3} & \smallincreasebg{} & \smallincreasebg{51.4} & \smallincreasebg{52.2} & \smallincreasebg{} & \smallincreasebg{54.1} & \smallincreasebg{56.7} & \smallincreasebg{} & \smallincreasebg{56.8} & \smallincreasebg{44.4} & \smallincreasebg{} & \smallincreasebg{52.8} & \smallincreasebg{52.2}  & \smallincreasebg{} & \smallincreasebg{68.1} & \smallincreasebg{75.6} & \smallincreasebg{} & \smallincreasebg{56.8} & \smallincreasebg{53.3} & \smallincreasebg{} & \smallincreasebg{62.9} & \smallincreasebg{68.9} & \smallincreasebg{} & \smallincreasebg{58.2} & \smallincreasebg{67.8} & \smallincreasebg{} & \smallincreasebg{57.1} & \smallincreasebg{58.3}\\
& Exact-Flan-T5* & 60.8 & 60.3 & & 55.7 & \textbf{56.9} &  &  61.3 & 59.2 & &   57.6 & 55.2 & & 58.1 & 57.8 & &  67.2 & 73.3 & &   57.1 & 54.3 & & 63.9 & 74.9 & & 63.0 & \textbf{73.9} & & 60.5 & 62.9\\
& Flan-T5* & \smallincreasebg{\textbf{63.6}} & \smallincreasebg{\textbf{62.2}} & \smallincreasebg{} & \smallincreasebg{\textbf{56.9}} & \smallincreasebg{55.6} & \smallincreasebg{} & \smallincreasebg{\textbf{62.9}} & \smallincreasebg{\textbf{61.1}} & \smallincreasebg{} & \smallincreasebg{\textbf{59.7}} & \smallincreasebg{\textbf{57.8}} & \smallincreasebg{}  &  \smallincreasebg{\textbf{60.8}} &  \smallincreasebg{\textbf{61.1}} & \smallincreasebg{} &  \smallincreasebg{\textbf{69.7}}  & \smallincreasebg{\textbf{76.7}} & \smallincreasebg{} &  \smallincreasebg{\textbf{60.4}}  & \smallincreasebg{\textbf{56.7}} & \smallincreasebg{} & \smallincreasebg{\textbf{64.3}}  & \smallincreasebg{\textbf{75.6}} & \smallincreasebg{} & \smallincreasebg{\textbf{63.6}}  & \smallincreasebg{72.2} & \smallincreasebg{} & \smallincreasebg{\textbf{62.4}} & \smallincreasebg{\textbf{64.3}}\\

\hline
\end{tabular}}
\end{adjustbox}
\caption{Supervised experimental results of \acs{QR} using \OurDatarank.}
\label{exp2_gpt4}
\vspace{-3mm}
\end{table*}

\section{Experiments}
\subsection{Experimental setup} 
\header{Dataset} 
We perform experiments on \acs{AG} and \acs{QR}. For each task, we report the single/cross-market results on the whole dataset and its subset. 

For \acs{AG}, on the \OurData dataset, we first adopt the BERT classifier trained in \cite{Gupta2019AmazonQAAR}. It assesses each question based on the review information, categorizing them as either answerable or unanswerable. Subsequently, we employ it to filter out all answerable questions. We then split the training/validation/testing sets following the portion of 70/10/20\%, resulting in 183,092/24,973/49,958 samples, respectively. On the \OurDataans dataset, we also split the data into three sets with the same portions. Specifically, we adopt the GPT-4 generated answers as the ground truth. In the single-market setting, we retrieve the top $K$ reviews from the main marketplace before generating the answers\footnote{We choose $K=5$ in our case.}. In the cross-market setting, we retrieve the reviews from both the main and auxiliary marketplaces. We report the generation performance of baselines on the testing set. 

For \acs{QR}, we first rank products, then among the top $N$ products, we rank the top $K$ questions\footnote{Following \cite{Ghasemi2023CrossMarketPQ}, we use $N=3$ and $K=50$.}. Since \OurData does not come with any ground-truth ranking results, we perform unsupervised training and adopt GPT-4-labeled data, \OurDatarank, as the testing set. Besides, to further test the performance of supervised methods on this task, we split \OurDatarank into three sets, with 1260/180/360 samples in each. We then train each model on the training set and report results on the testing set. 

% \subsection{Evaluation metrics}
\header{Evaluation metrics}
We adopt several evaluation metrics to assess the performance of models on two tasks. For \acs{AG}, we compare the model-generated answers with ground-truth user answers using BLEU-4~\cite{Papineni2002BleuAM} and ROUGE-L~\cite{Lin2004ROUGEAP} scores. For \acs{QR}, we report major information retrieval (IR) metrics, namely, mean reciprocal rank (MRR) and Precision@3 to evaluate the ranking performance of different methods.

\subsection{Compared methods}
% We adopt several baselines and assess their performance on our dataset on two tasks. 
For \acs{AG}, we first directly rank and select a review as the answer with methods such as BM25~\cite{Robertson2009ThePR}, BERT~\cite{Devlin2019BERTPO}. Besides, several generative methods such as T5~\cite{Raffel2019ExploringTL}, LLaMA-2~\cite{Touvron2023Llama2O}, are leveraged to train the model to generate the answer given the question and reviews. Specifically, under the cross-market scenario, Exact-model means that in the auxiliary marketplace, we only use reviews from the same item before performing answer generation.

For \acs{QR}, on \OurData, we report ranking methods that do not involve any training (\ie BERT, UPR~\cite{Sachan2022ImprovingPR}) or methods that perform unsupervised training (\ie CMJim~\cite{Ghasemi2023CrossMarketPQ}). On \OurDatarank, we adopt supervised fine-tuning methods (\ie BERT-f/monoT5~\cite{Nogueira2020DocumentRW}), and report testing performance. 
Details of each method are listed in Appendix~\ref{sec:baseline_details}.

\subsection{Experimental results}
\subsubsection{Review-based answer generation}
\label{exp:RBG}
Tables~\ref{exp1} and \ref{exp1_gpt4} show the single/cross-market answer generation performance on \OurData and \OurDataans datasets. We have the following observations: first of all, cross-market models have superior overall performance in all marketplaces compared with methods in the single-market setting. This result verifies \textbf{RQ1} from the model perspective, showing that external resources (\ie reviews), from auxiliary marketplaces, can significantly contribute to  improved outcomes in the main marketplace. A clear advantage of LLMs over traditional methods is evident across various marketplaces. Notably, LLaMA-2 outperforms the overall cross-market \OurData dataset, with a notable ROUGE improvement from 13.5 in T5 to 18.3. Similarly, in \OurDataans, the overall ROUGE score sees significant enhancement, rising from 29.1 to 38.5. This provides an answer for \textbf{RQ3}, offering insights into the efficacy and potential advancements of LLMs.
% in the MCPQA task.

\subsubsection{Product-related question ranking}
Tables~\ref{exp2} and \ref{exp2_gpt4} show the question ranking results within the single/cross-market scenario on two datasets. We notice that most observations from Section~\ref{exp:RBG} still hold. For example, performance advantages persist in product-related question ranking compared to a single-market scenario. This shows that a large number of relevant questions in the auxiliary marketplaces help address similar questions in a low-resource marketplace. Furthermore, the performance boost is more obvious in marketplaces with a smaller scale (\ie \textbf{au}, \textbf{br}) compared with marketplaces with a larger scale (\ie \textbf{uk}). For instance, the P@3 BM25 performance  exhibits an improvement 28.3 and 21.7 for \textbf{au} and \textbf{br} marketplaces, respectively, compared with 17.3 in \textbf{uk} on \OurData. We also find that in the cross-market setting, the Exact-models have a weaker overall performance than their original counterparts (\ie Exact-T5/Llama-2 v.s.\ T5/Llama-2). For example, on \OurDatarank, the cross-market Exact-Flan-T5 is 1.4 weaker in terms of overall P@3 compared with Flan-T5. This demonstrates that valuable information can be found within similar products from auxiliary marketplaces, even when they possess slightly different titles. We list some cases in Appendix \ref{case_study} to elaborate on this.

\begin{figure}[t]
\begin{subfigure}{0.45\columnwidth}
\includegraphics[width=\textwidth]{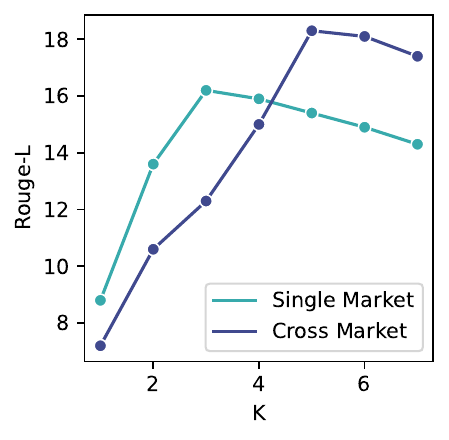}
\caption{\OurData}
\end{subfigure}
\begin{subfigure}{0.45\columnwidth}
\includegraphics[width=\textwidth]{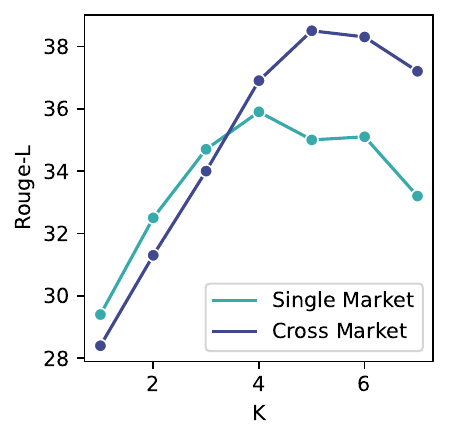}
\caption{\OurDataans}
\end{subfigure}

\begin{subfigure}{0.45\columnwidth}
\includegraphics[width=\textwidth]{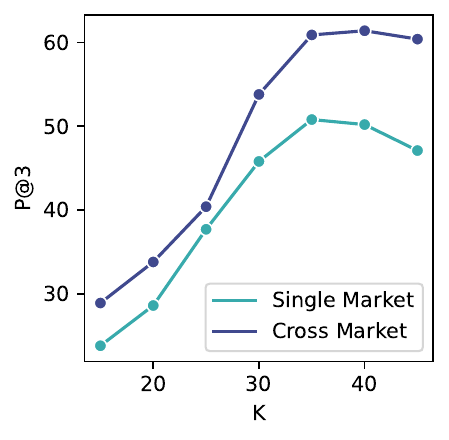}
\caption{\OurData}
\end{subfigure}
\begin{subfigure}{0.45\columnwidth}
\includegraphics[width=\textwidth]{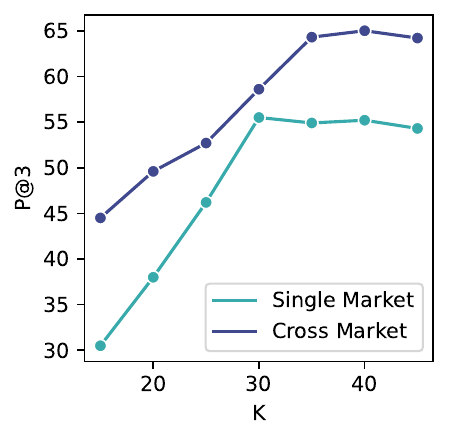}
\caption{\OurDatarank}
\end{subfigure}
\caption{$K$-value analysis across different marketplaces on the best-performed model. The upper row is on \acs{AG}, and the lower is \acs{QR}.}
\label{kanalysis}
\vspace{-3mm}
\end{figure}

\begin{figure}[t]

\begin{subfigure}{0.48\columnwidth}
\includegraphics[width=\textwidth]{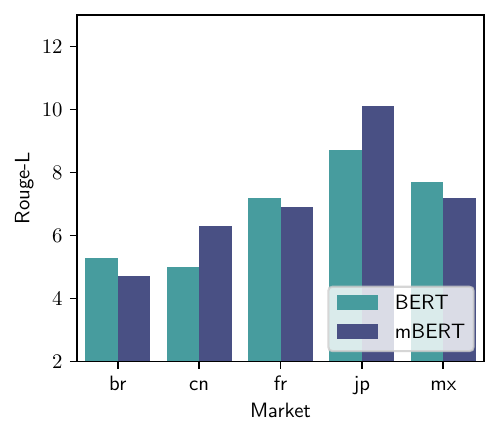}
\caption{\OurData}
\end{subfigure}
\begin{subfigure}{0.48\columnwidth}
\includegraphics[width=\textwidth]{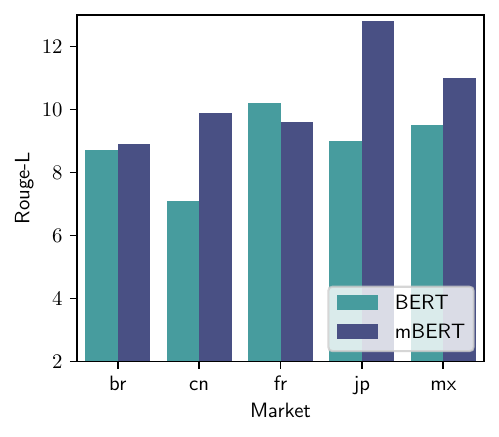}
\caption{\OurDataans}
\end{subfigure}

\begin{subfigure}{0.48\columnwidth}
\includegraphics[width=\textwidth]{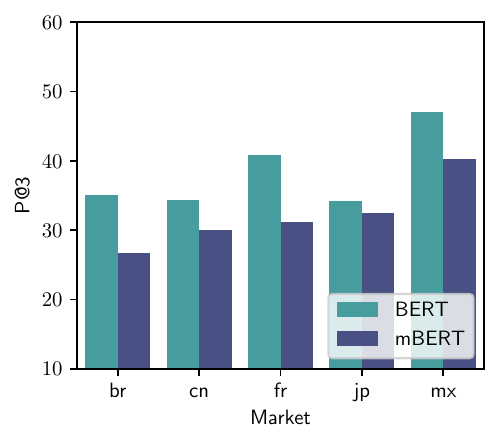}
\caption{\OurData}
\end{subfigure}
\begin{subfigure}{0.48\columnwidth}
\includegraphics[width=\textwidth]{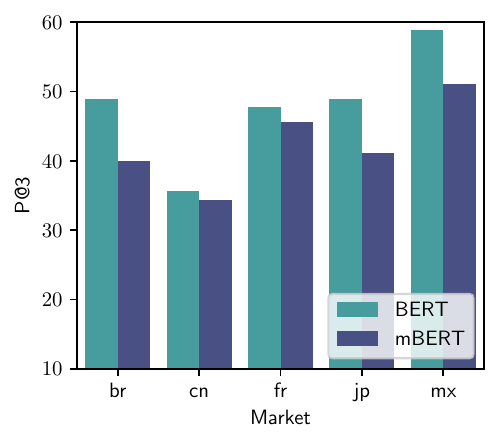}
\caption{\OurDatarank}
\end{subfigure}
\caption{Multilingual analysis on non-English marketplaces. The upper row is on \acs{AG}, the lower is \acs{QR}.}
\label{multilingualanalysis}
\vspace{-3mm}
\end{figure}

% \input{Tables/exp2_gpt4}
% \section{Potential tasks}
% \begin{itemize}
%     \item \textbf{Unsupervised Multilingual cross-market QA} --> We train the question embedding in an unsupervised way. For multilingual data, we can both use translation (can try LLMs \textit{i.e.} Llama) or use multilingual models like mBert. Can do both \verb|item-aware| and \verb|item-unaware|.
%     \item \textbf{Supervised Multilingual cross-market QA} --> For a question q, we first retrieve relevant questions, then we adopt ChatGPT to automatically label the similarity label to other questions. Or we do a small portion of human annotation, then we fine-tune llama to generate weakly supervised labels.
%     \item \textbf{Multilingual item ranking} --> Given an item find its relevant items.

% \end{itemize}

\section{External Analysis}
\subsection{Hyperparameter analysis}
We investigate the effect of the number of retrieved product-related resources (\ie questions, reviews) $K$ under both single/cross-market scenarios. We report the average performance among every marketplace on both \OurData and the corresponding subset in Figure \ref{kanalysis}. We observe that in \ac{AG}, initially, the performance of Llama-2 in the cross-market setting is inferior to that in the single market. However, after increasing the value of $K$, the optimal $K$ value in the cross-market scenario surpasses that in the single market. This tendency indicates that richer information is contained in the cross-market reviews. In \acs{QR}, the ranking performance in the single-market scenario begins to decline when $K$ is around 50. This indicates that some less relevant questions are retrieved, negatively impacting the results. Conversely, in the cross-market scenario, a greater number of relevant questions are accessible, helping to effectively mitigate this issue.

\subsection{Multilingual analysis}
We undertake a comparative analysis between translated and non-translated content to delve deeper into performance variations across non-English marketplaces. In particular, within the single-market scenario, we compare mBERT with BERT in 5 non-English marketplaces. Here, mBERT refers to a setup where all contents and the model itself are preserved and fine-tuned in their original language without  translation. The results are shown in Figure \ref{multilingualanalysis}.
We notice that in the \acs{AG} task, concerning some non-Latin languages (\ie \textbf{cn}, \textbf{jp}), the performance of single-market mBERT without translation results in higher score compared with BERT on two datasets. However, we observe opposite results in some other non-English marketplaces (\ie \textbf{fr}). Besides, in the \acs{QR} task, the performance of mBERT is inferior to the translated BERT model. This underscores a crucial future direction for this task: effectively enhancing performance in non-English marketplaces, an aspect that has been relatively underexplored. 
% Moreover, investigating the impact of domain adaptation and cross-lingual transfer learning techniques may offer insights into enhancing the robustness of models across diverse linguistic contexts.

% Model performance in terms of external marketsize?
% Or some analysis concerning multilingual? We notice that directly performing answer generation in non-English languages without translation still remains challenging in this context.

\section{Conclusions}
We propose the task of Multilingual Cross-market Product-based Question Answering (MCPQA). We hypothesize that product-related information from a resource-rich marketplace can be leveraged to enhance the QA in a resource-scarce marketplace.  To facilitate the research, we then propose a large-scale dataset, covering over 7 million questions across 17 marketplaces and 11 languages. Additionally, we perform automatic translation for the Electronics category, labeling it as \OurData. We also provide LLM-labeled subsets on \OurData for each of the two tasks, namely \OurDataans and \OurDatarank. Specifically, we focus on two different tasks: \acs{AG} and \acs{QR}. We conduct experiments to compare the performance of models under single/cross-market scenarios on both datasets.

\section*{Limitations}
The task of PQA holds significant potential in improving user experiences on e-commerce platforms. However, there are several limitations and challenges associated. One major challenge is the quality and reliability of the information available for answering user questions. Even though we make sure all of the information comes from real user-generated data, the reviews and QA pairs might still contain biased or inaccurate information. Furthermore, language barriers and the availability of data in multiple languages add complexity to the task of product-related QA, particularly in cross-lingual scenarios. The limited availability of data in low-resource languages further exacerbates this challenge. To address them, continued research and development efforts are still under process which aim at improving data quality, handling language diversity, etc. We discuss it as our future work in Appendix \ref{sec:appendix_future}.
% Currently, our dataset is limited to only 6 languages. We are still expanding our dataset by including more marketplaces such as Germany, Italy, Spain, etc. Also, marketplaces in low-resource languages are worthy of exploring under this scenario. The wealth of information in the auxiliary marketplace can prove effective in addressing questions within these low-resource language marketplaces. Moreover, we're also seeking to adopt more advanced models in improving the performance of answer generation/ranking. Even though LLMs such as LLaMA-2 have already shown promising results on this task, information such as product metadata should be better leveraged to improve the model. \moh{let's talk about the broader impact of the task and its limitations, rather than issues with the data.}

\section*{Ethics Statement}
\label{ethics}
Our dataset is derived from the publicly available product question-answering dataset, XMarket~\cite{Bonab2021CrossMarketPR}. We adhere to the  policies throughout the creation and utilization of this dataset to ensure the protection of user privacy. When preparing the question-answering pairs, we strictly ensure that no personally identifiable information is exposed or utilized in any form during the processes. We prioritize user privacy and confidentiality to maintain the integrity and ethical standards of our dataset. We have licensed our data under CC0 1.0 DEED and will ask the users to sign an agreement such that the dataset will only be available for academic research purposes to further protect the users. 

% Entries for the entire Anthology, followed by custom entries
\bibliography{custom}

\begin{thebibliography}{48}
\expandafter\ifx\csname natexlab\endcsname\relax\def\natexlab#1{#1}\fi

\bibitem[{Abbasiantaeb et~al.(2023)Abbasiantaeb, Yuan, Kanoulas, and Aliannejadi}]{Abbasiantaeb2023LetTL}
Zahra Abbasiantaeb, Yifei Yuan, E.~Kanoulas, and Mohammad Aliannejadi. 2023.
\newblock \href {https://api.semanticscholar.org/CorpusID:265659354} {Let the llms talk: Simulating human-to-human conversational qa via zero-shot llm-to-llm interactions}.
\newblock \emph{Proceedings of the 17th ACM International Conference on Web Search and Data Mining}.

\bibitem[{Artetxe et~al.(2019)Artetxe, Ruder, and Yogatama}]{Artetxe2019OnTC}
Mikel Artetxe, Sebastian Ruder, and Dani Yogatama. 2019.
\newblock \href {https://api.semanticscholar.org/CorpusID:204901567} {On the cross-lingual transferability of monolingual representations}.
\newblock In \emph{Annual Meeting of the Association for Computational Linguistics}.

\bibitem[{Asai et~al.(2020)Asai, Kasai, Clark, Lee, Choi, and Hajishirzi}]{Asai2020XORQC}
Akari Asai, Jungo Kasai, J.~Clark, Kenton Lee, Eunsol Choi, and Hannaneh Hajishirzi. 2020.
\newblock \href {https://api.semanticscholar.org/CorpusID:225040672} {Xor qa: Cross-lingual open-retrieval question answering}.
\newblock In \emph{North American Chapter of the Association for Computational Linguistics}.

\bibitem[{Bjerva et~al.(2020)Bjerva, Bhutani, Golshan, Tan, and Augenstein}]{Bjerva2020SubjQAAD}
Johannes Bjerva, Nikita Bhutani, Behzad Golshan, Wang~Chiew Tan, and Isabelle Augenstein. 2020.
\newblock \href {https://api.semanticscholar.org/CorpusID:216642239} {Subjqa: A dataset for subjectivity and review comprehension}.
\newblock In \emph{Conference on Empirical Methods in Natural Language Processing}.

\bibitem[{Bonab et~al.(2021)Bonab, Aliannejadi, Vardasbi, Kanoulas, and Allan}]{Bonab2021CrossMarketPR}
Hamed Bonab, Mohammad Aliannejadi, Ali Vardasbi, E.~Kanoulas, and James Allan. 2021.
\newblock \href {https://api.semanticscholar.org/CorpusID:237484242} {Cross-market product recommendation}.
\newblock \emph{Proceedings of the 30th ACM International Conference on Information \& Knowledge Management}.

\bibitem[{Chen et~al.(2019)Chen, Li, Ji, Zhou, and Chen}]{wsdm19-pqa-chen}
Shiqian Chen, Chenliang Li, Feng Ji, Wei Zhou, and Haiqing Chen. 2019.
\newblock \href {https://doi.org/10.1145/3289600.3290971} {Review-driven answer generation for product-related questions in e-commerce}.
\newblock In \emph{Proceedings of the Twelfth {ACM} International Conference on Web Search and Data Mining, {WSDM} 2019, Melbourne, VIC, Australia, February 11-15, 2019}, pages 411--419. {ACM}.

\bibitem[{Chung et~al.(2022)Chung, Hou, Longpre, Zoph, Tay, Fedus, Li, Wang, Dehghani, Brahma, Webson, Gu, Dai, Suzgun, Chen, Chowdhery, Valter, Narang, Mishra, Yu, Zhao, Huang, Dai, Yu, Petrov, hsin Chi, Dean, Devlin, Roberts, Zhou, Le, and Wei}]{Chung2022ScalingIL}
Hyung~Won Chung, Le~Hou, S.~Longpre, Barret Zoph, Yi~Tay, William Fedus, Eric Li, Xuezhi Wang, Mostafa Dehghani, Siddhartha Brahma, Albert Webson, Shixiang~Shane Gu, Zhuyun Dai, Mirac Suzgun, Xinyun Chen, Aakanksha Chowdhery, Dasha Valter, Sharan Narang, Gaurav Mishra, Adams~Wei Yu, Vincent Zhao, Yanping Huang, Andrew~M. Dai, Hongkun Yu, Slav Petrov, Ed~Huai hsin Chi, Jeff Dean, Jacob Devlin, Adam Roberts, Denny Zhou, Quoc~V. Le, and Jason Wei. 2022.
\newblock \href {https://api.semanticscholar.org/CorpusID:253018554} {Scaling instruction-finetuned language models}.
\newblock \emph{ArXiv}, abs/2210.11416.

\bibitem[{Clark et~al.(2020)Clark, Choi, Collins, Garrette, Kwiatkowski, Nikolaev, and Palomaki}]{Clark2020TyDiQA}
J.~Clark, Eunsol Choi, Michael Collins, Dan Garrette, Tom Kwiatkowski, Vitaly Nikolaev, and Jennimaria Palomaki. 2020.
\newblock \href {https://api.semanticscholar.org/CorpusID:212657414} {Tydi qa: A benchmark for information-seeking question answering in typologically diverse languages}.
\newblock \emph{Transactions of the Association for Computational Linguistics}, 8:454--470.

\bibitem[{Deng et~al.(2022)Deng, Li, Zhang, Ding, and Lam}]{tois22-pqa}
Yang Deng, Yaliang Li, Wenxuan Zhang, Bolin Ding, and Wai Lam. 2022.
\newblock \href {https://doi.org/10.1145/3507782} {Toward personalized answer generation in e-commerce via multi-perspective preference modeling}.
\newblock \emph{{ACM} Trans. Inf. Syst.}, 40(4):87:1--87:28.

\bibitem[{Deng et~al.(2018)Deng, Shen, Yang, Li, Du, Fan, and Lei}]{coling18-crossdomain}
Yang Deng, Ying Shen, Min Yang, Yaliang Li, Nan Du, Wei Fan, and Kai Lei. 2018.
\newblock \href {https://aclanthology.org/C18-1279/} {Knowledge as {A} bridge: Improving cross-domain answer selection with external knowledge}.
\newblock In \emph{Proceedings of the 27th International Conference on Computational Linguistics, {COLING} 2018, Santa Fe, New Mexico, USA, August 20-26, 2018}, pages 3295--3305. Association for Computational Linguistics.

\bibitem[{Deng et~al.(2020)Deng, Zhang, and Lam}]{cikm20-pqa}
Yang Deng, Wenxuan Zhang, and Wai Lam. 2020.
\newblock \href {https://doi.org/10.1145/3340531.3411904} {Opinion-aware answer generation for review-driven question answering in e-commerce}.
\newblock In \emph{{CIKM} '20: The 29th {ACM} International Conference on Information and Knowledge Management, Virtual Event, Ireland, October 19-23, 2020}, pages 255--264. {ACM}.

\bibitem[{Deng et~al.(2023)Deng, Zhang, Yu, and Lam}]{Deng2023ProductQA}
Yang Deng, Wenxuan Zhang, Qian Yu, and Wai Lam. 2023.
\newblock \href {https://api.semanticscholar.org/CorpusID:256901084} {Product question answering in e-commerce: A survey}.
\newblock \emph{Proceedings of the 61st Annual Meeting of the Association for Computational Linguistics}.

\bibitem[{Devlin et~al.(2019)Devlin, Chang, Lee, and Toutanova}]{Devlin2019BERTPO}
Jacob Devlin, Ming-Wei Chang, Kenton Lee, and Kristina Toutanova. 2019.
\newblock \href {https://api.semanticscholar.org/CorpusID:52967399} {Bert: Pre-training of deep bidirectional transformers for language understanding}.
\newblock In \emph{North American Chapter of the Association for Computational Linguistics}.

\bibitem[{Feng et~al.(2021)Feng, Ren, Zhao, Sun, and Li}]{sigir21-pqa}
Yue Feng, Zhaochun Ren, Weijie Zhao, Mingming Sun, and Ping Li. 2021.
\newblock \href {https://doi.org/10.1145/3404835.3462899} {Multi-type textual reasoning for product-aware answer generation}.
\newblock In \emph{{SIGIR} '21: The 44th International {ACM} {SIGIR} Conference on Research and Development in Information Retrieval, Virtual Event, Canada, July 11-15, 2021}, pages 1135--1145. {ACM}.

\bibitem[{Gao et~al.(2021)Gao, Chen, Ren, Zhao, and Yan}]{tois21-pqa}
Shen Gao, Xiuying Chen, Zhaochun Ren, Dongyan Zhao, and Rui Yan. 2021.
\newblock \href {https://doi.org/10.1145/3432689} {Meaningful answer generation of e-commerce question-answering}.
\newblock \emph{{ACM} Trans. Inf. Syst.}, 39(2):18:1--18:26.

\bibitem[{Gao et~al.(2019)Gao, Ren, Zhao, Zhao, Yin, and Yan}]{wsdm19-pqa}
Shen Gao, Zhaochun Ren, Yihong~Eric Zhao, Dongyan Zhao, Dawei Yin, and Rui Yan. 2019.
\newblock \href {https://doi.org/10.1145/3289600.3290992} {Product-aware answer generation in e-commerce question-answering}.
\newblock In \emph{Proceedings of the Twelfth {ACM} International Conference on Web Search and Data Mining, {WSDM} 2019, Melbourne, VIC, Australia, February 11-15, 2019}, pages 429--437. {ACM}.

\bibitem[{Ghasemi et~al.(2023)Ghasemi, Aliannejadi, Bonab, Kanoulas, de~Vries, Allan, and Hiemstra}]{Ghasemi2023CrossMarketPQ}
Negin Ghasemi, Mohammad Aliannejadi, Hamed Bonab, E.~Kanoulas, Arjen~P. de~Vries, James Allan, and Djoerd Hiemstra. 2023.
\newblock \href {https://api.semanticscholar.org/CorpusID:259950015} {Cross-market product-related question answering}.
\newblock \emph{Proceedings of the 46th International ACM SIGIR Conference on Research and Development in Information Retrieval}.

\bibitem[{Gonz{\'a}lez and G{\'o}mez(2007)}]{Gonzlez2007TRECEA}
Jos{\'e} Luis~Vicedo Gonz{\'a}lez and Jaime G{\'o}mez. 2007.
\newblock \href {https://api.semanticscholar.org/CorpusID:33367421} {Trec: Experiment and evaluation in information retrieval}.
\newblock \emph{J. Assoc. Inf. Sci. Technol.}, 58:910--911.

\bibitem[{Gupta et~al.(2019)Gupta, Kulkarni, Chanda, Rayasam, and Lipton}]{Gupta2019AmazonQAAR}
Mansi Gupta, Nitish Kulkarni, Raghuveer Chanda, Anirudha Rayasam, and Zachary~Chase Lipton. 2019.
\newblock \href {https://api.semanticscholar.org/CorpusID:199465954} {Amazonqa: A review-based question answering task}.
\newblock In \emph{International Joint Conference on Artificial Intelligence}.

\bibitem[{Lin(2004)}]{Lin2004ROUGEAP}
Chin-Yew Lin. 2004.
\newblock \href {https://api.semanticscholar.org/CorpusID:964287} {Rouge: A package for automatic evaluation of summaries}.
\newblock In \emph{Annual Meeting of the Association for Computational Linguistics}.

\bibitem[{Liu et~al.(2019)Liu, Lin, Liu, and Sun}]{Liu2019XQAAC}
Jiahua Liu, Yankai Lin, Zhiyuan Liu, and Maosong Sun. 2019.
\newblock \href {https://api.semanticscholar.org/CorpusID:196174566} {Xqa: A cross-lingual open-domain question answering dataset}.
\newblock In \emph{Annual Meeting of the Association for Computational Linguistics}.

\bibitem[{Longpre et~al.(2020)Longpre, Lu, and Daiber}]{Longpre2020MKQAAL}
S.~Longpre, Yi~Lu, and Joachim Daiber. 2020.
\newblock \href {https://api.semanticscholar.org/CorpusID:220871404} {Mkqa: A linguistically diverse benchmark for multilingual open domain question answering}.
\newblock \emph{Transactions of the Association for Computational Linguistics}, 9:1389--1406.

\bibitem[{McAuley and Yang(2015)}]{McAuley2015AddressingCA}
Julian McAuley and Alex Yang. 2015.
\newblock \href {https://api.semanticscholar.org/CorpusID:3690214} {Addressing complex and subjective product-related queries with customer reviews}.
\newblock \emph{Proceedings of the 25th International Conference on World Wide Web}.

\bibitem[{Nogueira et~al.(2020)Nogueira, Jiang, Pradeep, and Lin}]{Nogueira2020DocumentRW}
Rodrigo Nogueira, Zhiying Jiang, Ronak Pradeep, and Jimmy~J. Lin. 2020.
\newblock \href {https://api.semanticscholar.org/CorpusID:212725651} {Document ranking with a pretrained sequence-to-sequence model}.
\newblock In \emph{Findings}.

\bibitem[{OpenAI(2023)}]{Achiam2023GPT4TR}
OpenAI. 2023.
\newblock \href {https://api.semanticscholar.org/CorpusID:257532815} {Gpt-4 technical report}.

\bibitem[{Papineni et~al.(2002)Papineni, Roukos, Ward, and Zhu}]{Papineni2002BleuAM}
Kishore Papineni, Salim Roukos, Todd Ward, and Wei-Jing Zhu. 2002.
\newblock \href {https://api.semanticscholar.org/CorpusID:11080756} {Bleu: a method for automatic evaluation of machine translation}.
\newblock In \emph{Annual Meeting of the Association for Computational Linguistics}.

\bibitem[{Qu et~al.(2020)Qu, Ding, Liu, Liu, Ren, Zhao, Dong, Wu, and Wang}]{Qu2020RocketQAAO}
Yingqi Qu, Yuchen Ding, Jing Liu, Kai Liu, Ruiyang Ren, Xin Zhao, Daxiang Dong, Hua Wu, and Haifeng Wang. 2020.
\newblock \href {https://api.semanticscholar.org/CorpusID:231815627} {Rocketqa: An optimized training approach to dense passage retrieval for open-domain question answering}.
\newblock In \emph{North American Chapter of the Association for Computational Linguistics}.

\bibitem[{Raffel et~al.(2019)Raffel, Shazeer, Roberts, Lee, Narang, Matena, Zhou, Li, and Liu}]{Raffel2019ExploringTL}
Colin Raffel, Noam~M. Shazeer, Adam Roberts, Katherine Lee, Sharan Narang, Michael Matena, Yanqi Zhou, Wei Li, and Peter~J. Liu. 2019.
\newblock \href {https://api.semanticscholar.org/CorpusID:204838007} {Exploring the limits of transfer learning with a unified text-to-text transformer}.
\newblock \emph{J. Mach. Learn. Res.}, 21:140:1--140:67.

\bibitem[{Robertson and Zaragoza(2009)}]{Robertson2009ThePR}
Stephen~E. Robertson and Hugo Zaragoza. 2009.
\newblock \href {https://api.semanticscholar.org/CorpusID:207178704} {The probabilistic relevance framework: Bm25 and beyond}.
\newblock \emph{Found. Trends Inf. Retr.}, 3:333--389.

\bibitem[{Rozen et~al.(2021)Rozen, Carmel, Mejer, Mirkis, and Ziser}]{naacl21-pqa}
Ohad Rozen, David Carmel, Avihai Mejer, Vitaly Mirkis, and Yftah Ziser. 2021.
\newblock \href {https://doi.org/10.18653/v1/2021.naacl-main.23} {Answering product-questions by utilizing questions from other contextually similar products}.
\newblock In \emph{Proceedings of the 2021 Conference of the North American Chapter of the Association for Computational Linguistics: Human Language Technologies, {NAACL-HLT} 2021, Online, June 6-11, 2021}, pages 242--253. Association for Computational Linguistics.

\bibitem[{Sachan et~al.(2022)Sachan, Lewis, Joshi, Aghajanyan, tau Yih, Pineau, and Zettlemoyer}]{Sachan2022ImprovingPR}
Devendra~Singh Sachan, Mike Lewis, Mandar Joshi, Armen Aghajanyan, Wen tau Yih, Jo{\"e}lle Pineau, and Luke Zettlemoyer. 2022.
\newblock \href {https://api.semanticscholar.org/CorpusID:248218489} {Improving passage retrieval with zero-shot question generation}.
\newblock In \emph{Conference on Empirical Methods in Natural Language Processing}.

\bibitem[{Shen et~al.(2023)Shen, Asai, Byrne, and Gispert}]{Shen2023xPQACP}
Xiaoyu Shen, Akari Asai, Bill Byrne, and A.~Gispert. 2023.
\newblock \href {https://api.semanticscholar.org/CorpusID:258715321} {xpqa: Cross-lingual product question answering in 12 languages}.
\newblock \emph{Proceedings of the 61st Annual Meeting of the Association for Computational Linguistics}.

\bibitem[{Shen et~al.(2022)Shen, Barlacchi, Tredici, Cheng, and Gispert}]{Shen2022semiPQAAS}
Xiaoyu Shen, Gianni Barlacchi, Marco~Del Tredici, Weiwei Cheng, and A.~Gispert. 2022.
\newblock \href {https://api.semanticscholar.org/CorpusID:248779912} {semipqa: A study on product question answering over semi-structured data}.
\newblock \emph{Proceedings of The Fifth Workshop on e-Commerce and NLP (ECNLP 5)}.

\bibitem[{team et~al.(2022)team, Costa-juss{\`a}, Cross, cCelebi, Elbayad, Heafield, Heffernan, Kalbassi, Lam, Licht, Maillard, Sun, Wang, Wenzek, Youngblood, Akula, Barrault, Gonzalez, Hansanti, Hoffman, Jarrett, Sadagopan, Rowe, Spruit, Tran, Andrews, Ayan, Bhosale, Edunov, Fan, Gao, Goswami, Guzm'an, Koehn, Mourachko, Ropers, Saleem, Schwenk, and Wang}]{team2022NoLL}
Nllb team, Marta~Ruiz Costa-juss{\`a}, James Cross, Onur cCelebi, Maha Elbayad, Kenneth Heafield, Kevin Heffernan, Elahe Kalbassi, Janice Lam, Daniel Licht, Jean Maillard, Anna Sun, Skyler Wang, Guillaume Wenzek, Alison Youngblood, Bapi Akula, Lo{\"i}c Barrault, Gabriel~Mejia Gonzalez, Prangthip Hansanti, John Hoffman, Semarley Jarrett, Kaushik~Ram Sadagopan, Dirk Rowe, Shannon~L. Spruit, C.~Tran, Pierre~Yves Andrews, Necip~Fazil Ayan, Shruti Bhosale, Sergey Edunov, Angela Fan, Cynthia Gao, Vedanuj Goswami, Francisco Guzm'an, Philipp Koehn, Alexandre Mourachko, Christophe Ropers, Safiyyah Saleem, Holger Schwenk, and Jeff Wang. 2022.
\newblock \href {https://api.semanticscholar.org/CorpusID:250425961} {No language left behind: Scaling human-centered machine translation}.
\newblock \emph{ArXiv}, abs/2207.04672.

\bibitem[{Touvron et~al.(2023{\natexlab{a}})Touvron, Lavril, Izacard, Martinet, Lachaux, Lacroix, Rozi{\`e}re, Goyal, Hambro, Azhar, Rodriguez, Joulin, Grave, and Lample}]{Touvron2023LLaMAOA}
Hugo Touvron, Thibaut Lavril, Gautier Izacard, Xavier Martinet, Marie-Anne Lachaux, Timoth{\'e}e Lacroix, Baptiste Rozi{\`e}re, Naman Goyal, Eric Hambro, Faisal Azhar, Aurelien Rodriguez, Armand Joulin, Edouard Grave, and Guillaume Lample. 2023{\natexlab{a}}.
\newblock \href {https://api.semanticscholar.org/CorpusID:257219404} {Llama: Open and efficient foundation language models}.
\newblock \emph{ArXiv}, abs/2302.13971.

\bibitem[{Touvron et~al.(2023{\natexlab{b}})Touvron, Martin, Stone, Albert, Almahairi, Babaei, Bashlykov, Batra, Bhargava, Bhosale, Bikel, Blecher, Ferrer, Chen, Cucurull, Esiobu, Fernandes, Fu, Fu, Fuller, Gao, Goswami, Goyal, Hartshorn, Hosseini, Hou, Inan, Kardas, Kerkez, Khabsa, Kloumann, Korenev, Koura, Lachaux, Lavril, Lee, Liskovich, Lu, Mao, Martinet, Mihaylov, Mishra, Molybog, Nie, Poulton, Reizenstein, Rungta, Saladi, Schelten, Silva, Smith, Subramanian, Tan, Tang, Taylor, Williams, Kuan, Xu, Yan, Zarov, Zhang, Fan, Kambadur, Narang, Rodriguez, Stojnic, Edunov, and Scialom}]{Touvron2023Llama2O}
Hugo Touvron, Louis Martin, Kevin~R. Stone, Peter Albert, Amjad Almahairi, Yasmine Babaei, Nikolay Bashlykov, Soumya Batra, Prajjwal Bhargava, Shruti Bhosale, Daniel~M. Bikel, Lukas Blecher, Cristian~Cant{\'o}n Ferrer, Moya Chen, Guillem Cucurull, David Esiobu, Jude Fernandes, Jeremy Fu, Wenyin Fu, Brian Fuller, Cynthia Gao, Vedanuj Goswami, Naman Goyal, Anthony~S. Hartshorn, Saghar Hosseini, Rui Hou, Hakan Inan, Marcin Kardas, Viktor Kerkez, Madian Khabsa, Isabel~M. Kloumann, A.~V. Korenev, Punit~Singh Koura, Marie-Anne Lachaux, Thibaut Lavril, Jenya Lee, Diana Liskovich, Yinghai Lu, Yuning Mao, Xavier Martinet, Todor Mihaylov, Pushkar Mishra, Igor Molybog, Yixin Nie, Andrew Poulton, Jeremy Reizenstein, Rashi Rungta, Kalyan Saladi, Alan Schelten, Ruan Silva, Eric~Michael Smith, R.~Subramanian, Xia Tan, Binh Tang, Ross Taylor, Adina Williams, Jian~Xiang Kuan, Puxin Xu, Zhengxu Yan, Iliyan Zarov, Yuchen Zhang, Angela Fan, Melanie Kambadur, Sharan Narang, Aurelien Rodriguez, Robert Stojnic, Sergey Edunov, and
  Thomas Scialom. 2023{\natexlab{b}}.
\newblock \href {https://api.semanticscholar.org/CorpusID:259950998} {Llama 2: Open foundation and fine-tuned chat models}.
\newblock \emph{ArXiv}, abs/2307.09288.

\bibitem[{Wan and McAuley(2016)}]{Wan2016ModelingAS}
Mengting Wan and Julian McAuley. 2016.
\newblock \href {https://api.semanticscholar.org/CorpusID:6450583} {Modeling ambiguity, subjectivity, and diverging viewpoints in opinion question answering systems}.
\newblock \emph{2016 IEEE 16th International Conference on Data Mining (ICDM)}, pages 489--498.

\bibitem[{Xu et~al.(2019)Xu, Liu, Shu, and Yu}]{Xu2019BERTPF}
Hu~Xu, Bing Liu, Lei Shu, and Philip~S. Yu. 2019.
\newblock \href {https://api.semanticscholar.org/CorpusID:102353837} {Bert post-training for review reading comprehension and aspect-based sentiment analysis}.
\newblock In \emph{North American Chapter of the Association for Computational Linguistics}.

\bibitem[{Yu et~al.(2017)Yu, Qiu, Jiang, Huang, Song, Ji, Chu, and Chen}]{Yu2017ModellingDR}
Jianfei Yu, Minghui Qiu, Jing Jiang, Jun Huang, Shuangyong Song, Feng Ji, Wei Chu, and Haiqing Chen. 2017.
\newblock \href {https://api.semanticscholar.org/CorpusID:13314839} {Modelling domain relationships for transfer learning on retrieval-based question answering systems in e-commerce}.
\newblock \emph{Proceedings of the Eleventh ACM International Conference on Web Search and Data Mining}.

\bibitem[{Yu et~al.(2012)Yu, Zha, and Chua}]{emnlp12-pqa}
Jianxing Yu, Zheng{-}Jun Zha, and Tat{-}Seng Chua. 2012.
\newblock \href {https://aclanthology.org/D12-1036/} {Answering opinion questions on products by exploiting hierarchical organization of consumer reviews}.
\newblock In \emph{Proceedings of the 2012 Joint Conference on Empirical Methods in Natural Language Processing and Computational Natural Language Learning, EMNLP-CoNLL 2012, July 12-14, 2012, Jeju Island, Korea}, pages 391--401. {ACL}.

\bibitem[{Yu and Lam(2018)}]{Yu2018ReviewAwareAP}
Qian Yu and Wai Lam. 2018.
\newblock \href {https://api.semanticscholar.org/CorpusID:44884903} {Review-aware answer prediction for product-related questions incorporating aspects}.
\newblock \emph{Proceedings of the Eleventh ACM International Conference on Web Search and Data Mining}.

\bibitem[{Yuan and Lam(2021)}]{Yuan2021ConversationalFI}
Yifei Yuan and Wai Lam. 2021.
\newblock \href {https://api.semanticscholar.org/CorpusID:235368051} {Conversational fashion image retrieval via multiturn natural language feedback}.
\newblock \emph{Proceedings of the 44th International ACM SIGIR Conference on Research and Development in Information Retrieval}.

\bibitem[{Zhang et~al.(2019{\natexlab{a}})Zhang, Westerfield, Shim, Bingham, Fabbri, Verma, Hu, and Radev}]{Zhang2019ImprovingLC}
Rui Zhang, Caitlin Westerfield, Sungrok Shim, Garrett Bingham, Alexander~R. Fabbri, Neha Verma, William Hu, and Dragomir~R. Radev. 2019{\natexlab{a}}.
\newblock \href {https://api.semanticscholar.org/CorpusID:182952611} {Improving low-resource cross-lingual document retrieval by reranking with deep bilingual representations}.
\newblock \emph{ArXiv}, abs/1906.03492.

\bibitem[{Zhang et~al.(2019{\natexlab{b}})Zhang, Lau, Zhang, Chan, and Paris}]{Zhang2019DiscoveringRR}
Shiwei Zhang, Jey~Han Lau, Xiuzhen Zhang, Jeffrey Chan, and C{\'e}cile Paris. 2019{\natexlab{b}}.
\newblock \href {https://api.semanticscholar.org/CorpusID:210991889} {Discovering relevant reviews for answering product-related queries}.
\newblock \emph{2019 IEEE International Conference on Data Mining (ICDM)}, pages 1468--1473.

\bibitem[{Zhang et~al.(2020{\natexlab{a}})Zhang, Zhang, Lau, Chan, and Paris}]{pkdd20-pqa}
Shiwei Zhang, Xiuzhen Zhang, Jey~Han Lau, Jeffrey Chan, and C{\'{e}}cile Paris. 2020{\natexlab{a}}.
\newblock \href {https://doi.org/10.1007/978-3-030-67664-3\_34} {Less is more: Rejecting unreliable reviews for product question answering}.
\newblock In \emph{Machine Learning and Knowledge Discovery in Databases - European Conference, {ECML} {PKDD} 2020, Ghent, Belgium, September 14-18, 2020, Proceedings, Part {III}}, volume 12459 of \emph{Lecture Notes in Computer Science}, pages 567--583. Springer.

\bibitem[{Zhang et~al.(2020{\natexlab{b}})Zhang, Deng, and Lam}]{Zhang2020AnswerRF}
Wenxuan Zhang, Yang Deng, and Wai Lam. 2020{\natexlab{b}}.
\newblock \href {https://api.semanticscholar.org/CorpusID:220251219} {Answer ranking for product-related questions via multiple semantic relations modeling}.
\newblock \emph{Proceedings of the 43rd International ACM SIGIR Conference on Research and Development in Information Retrieval}.

\bibitem[{Zhang et~al.(2020{\natexlab{c}})Zhang, Deng, Ma, and Lam}]{emnlp20-pqa}
Wenxuan Zhang, Yang Deng, Jing Ma, and Wai Lam. 2020{\natexlab{c}}.
\newblock \href {https://doi.org/10.18653/v1/2020.emnlp-main.188} {Answerfact: Fact checking in product question answering}.
\newblock In \emph{Proceedings of the 2020 Conference on Empirical Methods in Natural Language Processing, {EMNLP} 2020, Online, November 16-20, 2020}, pages 2407--2417. Association for Computational Linguistics.

\bibitem[{Zhang et~al.(2020{\natexlab{d}})Zhang, Lam, Deng, and Ma}]{www20-pqa}
Wenxuan Zhang, Wai Lam, Yang Deng, and Jing Ma. 2020{\natexlab{d}}.
\newblock \href {https://doi.org/10.1145/3366423.3380015} {Review-guided helpful answer identification in e-commerce}.
\newblock In \emph{{WWW} '20: The Web Conference 2020, Taipei, Taiwan, April 20-24, 2020}, pages 2620--2626. {ACM} / {IW3C2}.

\end{thebibliography}
\bibliographystyle{acl_natbib}

\appendix

\section{LLM annotation details}
\label{sec:appendix}
We employ GPT-4 as the base LLM to perform automatic annotation. Specifically, \verb|gpt-4-1106-preview| is adopted in our setting. For review-based answer generation, we pass the question, related reviews into the model, and ask GPT-4 to generate if the corresponding answer can be produced from the given information and write the answer if possible. We also instruct GPT-4 to provide the corresponding reason. We use the following prompt:
\begin{itemize}
    \item \textit{In this task, you will be given a product question, and some reviews. You should judge if the reviews are helpful for answering the question. If yes, please write the corresponding answer and the reason. If no, please give the corresponding reason and provide the answer as no answer. Please output the answer format as: Judgement:yes/no, Reason: , Answer:} 
\end{itemize}

In our setup for product-related question ranking, we follow the annotation setting outlined in ~\cite{Ghasemi2023CrossMarketPQ}. Here, we utilize GPT-4 to evaluate the relevance of other question-answer pairs. The model is presented with two question-answer pairs from distinct products along with their respective product titles. Its task is to assess whether the QA pair associated with the second product proves useful in addressing the questions posed for the first product. Similarly, the model is also requested to provide the reason for making the judgment. The prompt is given as follows:
\begin{itemize}
    \item \textit{In this task, you will be given two different products, namely, Product A and B, respectively. Each product is associated with a question-answer pair. You should judge if the question-answer pair to Product B is useful for answering the question to Product A. You should assign a score from 0--2, as 0 represents not useful, 1 represents partially useful, and 2 represents very useful. Please also give the corresponding reason for making the decision. Please output the answer format as: Judgement:[score], Reason: }
\end{itemize}

\section{Future Directions}
\label{sec:appendix_future}
Future directions for the MCPQA task could involve several areas of exploration. First of all, more efforts could be put in the continued advancement and refinement of multilingual models capable of understanding and generating text across multiple languages. Furthermore, as a substantial portion of our dataset remains in its original, untranslated form, we are actively researching how models perform when fine-tuned on this untranslated data. Our focus lies particularly on assessing their question-answering performance in multilingual contexts. Based on that, investigation of cross-lingual transfer learning techniques to facilitate knowledge transfer and adaptation between languages could also be a promising direction in this task. This includes exploring approaches for transferring knowledge from high-resource to low-resource languages and vice versa.

\section{Human evaluation metrics}
\label{sec:human_metrics}
\begin{itemize}
    \item \textit{Correctness} aims to judge whether GPT-4 answers accurately serve as correct answers to the question, based on the given information. For example, if the question is not answerable from the reviews, GPT-4 should make the corresponding judgment. Otherwise, GPT-4 should first classify the question as answerable, and then give the corresponding answer.
    \item \textit{Completeness} is designed to determine whether the GPT-4 generated answers are complete and cover all aspects of the question.
    \item \textit{Relevance} is designed to determine whether the GPT-4 answers are relevant to the question, and whether contain hallucination that does not correspond to the original question.
    \item \textit{Naturalness} aims to determine whether the GPT-4 answers are smooth and natural. Whether there are obvious language errors and inconsistencies.
\end{itemize}
\begin{table*}[!h]
% \vspace{-1mm}
\small
    \centering
    \setlength{\tabcolsep}{1.5mm}
    \caption{Real cases of the single/cross-market question answering results. The first two are review-based generated answers. The rest shows answers obtained by product-related question ranking. The information in the bracket shows the retrieved related review/question. All the information is shown in translated English. }
    \begin{tabular}{cp{3.0cm} p{4.6cm} p{5.3cm} p{1.5cm}}
    \toprule
        \textbf{Idx} & \textbf{Question} & \textbf{Single-market ans.} & \textbf{Cross-market ans.}  &  \textbf{Gold ans.}\\
        \midrule
    1 &  Which Dell Inspiron N4010 model can I buy and be compatible with? & (The express delivery was very fast. It arrived the next day.) I have no idea what to answer. & (I bought this to double my RAM on my new Dell Inspiron 7559.) The Dell Inspiron 7559 model is compatible with this product. & This model is low voltage...\\
    2 & Sir is it compatible with x-T3? & (The picture quality is very sharp. Very sharp indeed.) The given review information cannot provide the answer. & (Bokeh is awesome quality of photos is amazing paired with the xt3) Yes, the lens is compatible with xt3. & Yes \\
     \midrule
    3 &  Will this work for a small fan that's 18 watts?  & (Will this work for charding an electric razor?) It provides power out the same as a wall power point.  So yes it should charge an electric razor.&(Will a travel fan work with this?) ... A travel fan should work this voltage converter. The converter has a maximum of 200 watts... & I think yes ... \\ %...    \textbf{market} The converter has a maximum of 200 watts and a small travel fan should be less than 50... Does it have an audio and hdmi output? <SEP> Good morning.\nIt has a micro HDMI output"
    4 & Does it have an external microphone input? & (Does it have an audio and hdmi output?) Good morning. It has a micro HDMI output. & (Can you put an external microphone?) No microphone input &  Doesn't have it! \\
\bottomrule
    \end{tabular} 
    \label{tab:casestudy}
\end{table*}

\section{Baseline details}
\label{sec:baseline_details}
We provide a detailed explanation of the baseline models we implement.

\header{Review-based answer generation} In this task, we report performance on \OurData and \OurDataans. In contrast to utilizing human answers in \OurData, in \OurDataans, we employ the GPT-4 generated results as the ground truth. For each dataset, we split the training/validation/testing set with the portion 70/10/20\% and report the results on the  testing set. The detailed information of each baseline is as follows:
\begin{itemize}
    \item BM25~\cite{Robertson2009ThePR} retrieves the top 5 reviews and adopts the top one directly as the answer.
    \item BERT~\cite{Devlin2019BERTPO} adopts a BERT ranker to re-rank the reviews retrieved by the top 100 BM25 results. Then the top 1 review is selected as the answer.
    % \item mBERT~\cite{Libovick2019HowLI} is similar to the Bert method except for a mBERT backbone is adopted where all the non-English contents are directly fed into the model without translation.
    \item T5~\cite{Raffel2019ExploringTL} takes the BM25 top 5 reviews as input and is fine-tuned to generate the corresponding answer.
    % \item mT5~\cite{Xue2020mT5AM} is fine-tuned in a similar setting as T5, with the exception that we utilize the mT5 backbone and abstain from using translated non-English contents.
    \item Exact-T5~\cite{Ghasemi2023CrossMarketPQ} is an answer generation model based on T5, wherein we initially identify the exact same item in the auxiliary marketplace and exclusively utilize the top 5 reviews among them as input.
    \item LLaMA-2~\cite{Touvron2023Llama2O} is in a similar setting as T5 but adopts LLaMA-2 as the backbone.
    \item Exact-LLaMA-2 is in a similar setting as Exact-T5 but adopts LLaMA-2 as the backbone.
    
\end{itemize}

\header{Product-related question ranking} In this task, we also report results on \OurData and \OurDatarank. Given that the \OurDatarank subset is the only portion in \OurData that contains ranking labels, Table~\ref{exp2} exclusively showcases unsupervised methods that leverage the remaining \OurData as the training set and subsequently present results on the \OurDatarank subset. Besides, to show the performance of supervised methods in this task, Table \ref{exp2_gpt4} splits \OurDatarank as the training/validation/testing set following the same portion as before. Performance is then reported on the testing set. 

We first provide details for the unsupervised methods in Table \ref{exp2}:
\begin{itemize}
    \item BM25~\cite{Robertson2009ThePR} reports the top-50 BM25 ranking results.
    \item BERT~\cite{Devlin2019BERTPO} performs BERT re-rank on BM25 top results.
    % \item mBert~\cite{Libovick2019HowLI} performs mBert re-rank on BM25 top-100 results without translation.
    \item UPR-m~\cite{Sachan2022ImprovingPR} is an unsupervised ranking method where we use a PLM to compute the probability of the input question conditioned on a related question. We use T5-base as the backbone.
    \item UPR-l adopts the same structure as UPR-m but uses T0-3B as the backbone.
    \item CMJim~\cite{Ghasemi2023CrossMarketPQ} is an unsupervised method that ranks products and their corresponding questions across marketplaces.
    \item Exact-\{BERT/UPR-l\} ranks the questions of the item from the main marketplace as well as the exact same item in the auxiliary marketplace.
\end{itemize}
\begin{table*}[!h]
% \vspace{-1mm}
\small
    \caption{Examples of data samples in \OurData.}
    \centering
    \setlength{\tabcolsep}{1.5mm}
    \begin{tabular}{cp{4.0cm} p{2.5cm} p{5.0cm} p{3.2cm}}
    \toprule
        \textbf{Market} & \textbf{Product title} & \textbf{Question} & \textbf{Reviews}& \textbf{Answer}    \\
        \midrule
        br & Sony - HDRCX405 HD Video Recording Handycam Camcorder (black) & É compatível com eos 80d?&Objetiva com desempenho muito bom. Estabilização de imagem (IS) funciona muito bem para uso sem tripé. STM com foco silencioso. Cumpre o que promete. & Bom dia, é totalmente compatível. \\
        cn & AKG Pro Audio K612 PRO Over-Ear, Open-Back, Premium Reference Studio Headphones & \begin{CJK*}{UTF8}{gbsn}akg品控真有那么差吗还是一群职业黑？\end{CJK*}  & \begin{CJK*}{UTF8}{gbsn}一言难尽。买了十几天刚煲开右耳时响时不响。现在退货中\end{CJK*} & \begin{CJK*}{UTF8}{gbsn}没有问题，还可以\end{CJK*}\\
        fr &  ViewSonic VG2439SMH 24 Inch 1080p Ergonomic Monitor with HDMI DisplayPort and VGA for Home and Office, Black & Sur écran webcam il y a t'il du son ? fait t'il webcam et micro en même temps?   &  Après réception; et déballage : produit simple et mise en marche facile. J'ai commandé deux écrans pour une station de travail. l’utilisateur est à l'aise & Pas le microphone. Webcam ok Son ok\\
        jp & SanDisk Ultra 64GB USB 3.0 OTG Flash Drive With micro USB connector For Android Mobile Devices(SDDD2-064G-G46) & \begin{CJK}{UTF8}{min}
            A1954に多用できますか\end{CJK} &  \begin{CJK}{UTF8}{min}小さすぎて使いにくい（笑）商品は、ゆうメールですぐに配達されました。\end{CJK} & \begin{CJK}{UTF8}{min}A1954とは、何ですか？キーボードは、英語配列です。\end{CJK}\\
        mx &  ZOTAC GeForce GT 730 1GB DDR3 PCI Express 2.0 x1 Graphics Card (ZT-71107-10L) & hola, es compatible con Lenovo TS-140? & Excelente producto y buen desempeño. Muy recomendable.& No conozco este equipo, solo se puede instalar en interfaces PIC x16.\\
        uk & Peachtree Audio Deepblue2 High Performance Wireless Bluetooth Music System (Black) & Can you play music through this speaker as a wired device from an mp3 player.&Reluctant to pay so much but couldn't be more happy. Amazing sound quality and worth every penny. You will be blown away. & Yes, with the supplied cable plugged into the headphone jack on the MP3 player and the auxiliary input on the deepblue2.\\

\bottomrule
    \end{tabular} 
    \label{tab:datasetexample1}
\end{table*}

We then detail the supervised methods in Table~\ref{exp2_gpt4}:
\begin{itemize}
    \item Bert-f~\cite{Devlin2019BERTPO} fine-tunes the Bert ranker on the training set.
    % For mBert, we perform training without translation.
    \item T5 is trained to generate the sequence of the ranked questions.
    \item monoT5~\cite{Nogueira2020DocumentRW} is another ranking method that takes T5 as backbone. We fine-tune the model on the training set and report the results on the testing portion. 
    % For mT5, we perform training without translation.
    \item Flan-T5~\cite{Chung2022ScalingIL} adopts the same structure as the monoT5 method but replaces the backbone to the Flan-T5-XL LLM.
    \item Exact-\{BERT-f/monoT5/Flan-T5\}~\cite{Ghasemi2023CrossMarketPQ} ranks the questions of the item from the main marketplace as well as the exact same item in the auxiliary marketplace.
\end{itemize}
\begin{table*}[!h]
% \vspace{-1mm}
\small
    \centering
    \setlength{\tabcolsep}{1.5mm}
    \begin{tabular}{cp{2.4cm}p{3.0cm} p{2,8cm} p{2.0cm} p{3.9cm}}
    \toprule
        \textbf{Market} & \textbf{Question}& \textbf{Reviews} & \textbf{GPT-4 ans.} & \textbf{Gold Ans.}& \textbf{GPT-4 reason}    \\
        \midrule
        au & Can you use this with Xbox one? & ... These however do not work as an Xbox one gamechat since there is no mic ...&No, you cannot use these headphones with Xbox One as there is no mic & Yes.  & Review 1 mentions that these headphones do not work as an Xbox One gamechat as there is no mic.\\
        ca & Will it work with the Nikon D3200 & I have the lens fixed on to my Nikon D3200 almost permanently...  & Yes, it will work with the Nikon D3200. & Works on my d3100. & The reviews mention
 that the product is compatible with the Nikon D3200 and they are using it effectively.  \\
        in & Does it has diopter correction ring? &  Extremely well built. Great lens and use it mostly for bird watching during my hikes... & No answer. & Yes. & None of the reviews mention a diopter correction ring for the product. \\
        jp & Does it come with a conversion plug to connect to an amplifier?  & ...They don't come with an adapter that can adjust to the standard plug, so be careful. & No, it does not come with a conversion plug to connect to an amplifier. & N/A. & Review 1 states explicitly that the headphones do not come with an adapter to adjust to the standard plug...      \\
        mx & How quiet is it? & ...Good performance, very quiet; Nice and quiet easily installed and works well...   &  The product is very quiet. & It is very quiet, reliable, highly recommended & Reviews 2, 3, 4, and 5 directly address the noise level of the product by stating it is `very quiet' and `nice and quiet' \\
        
\bottomrule
    \end{tabular} 
    \caption{Examples of data samples in \OurDataans. All the data is translated into English.}
    \label{tab:datasetexample2}
\end{table*}
\begin{table*}[!h]
% \vspace{-1mm}
\small

    \centering
    \setlength{\tabcolsep}{1.5mm}
    \begin{tabular}{cp{1.8cm}p{3.7cm} p{1.7cm} p{3.0cm} cp{3.2cm}}
    \toprule
        \textbf{Market} & \textbf{Product A}& \textbf{Product A QA} & \textbf{Product B} & \textbf{Product B QA}&  \textbf{tag} &\textbf{GPT-4 reason}    \\
         \midrule
        au &  Neewer 48 Macro LED Ring Flash Bundle with LCD Display Power Control... & Will this work with fuji x-t3 and x-t20? -> As long as they have a hot shoe, it will work. There is several lens ring adaptors for various lens sizes (talking about changeable lenses of course). & Neewer 48 Macro LED Ring Flash Bundle with LCD Display Power Control... & Is this compatible with FujifilmX-T3? -> As long as you have a hotshoe it should work. & 2 & Both Product A and Product B are the same Neewer 48 Macro LED Ring Flash Bundle, and the questions for both are concerning the compatibility with Fujifilm X-T3...\\
        cn & Kingston Digital Multi-Kit/Mobility Kit 16 GB ... & Hello, what is the writing speed of this micro sdxc? -> Write: 14Mo/s | Read: 20Mo/s ... & Kingston Digital Multi-Kit/Mobility Kit 16 GB...  & Speed of the card? -> Class 4 IE 4MB/sec. & 1 & The answer to Product B provides the class rating of a microSDHC card, though different from Product A... \\
        fr & iPad Air New iPad 9.7 inch 2017 Case... & Good evening, is this case compatible with an iPad 2? Thank you -> Yes, no problem. & iPad Air New iPad 9.7 inch 2017 Case...& Does  this case fit the ipad air 2? -> Hi, This case is not compatible with the iPad Air 2. & 0 & Product A is asking about iPad 2, while Product B is about compatibility with an iPad Air 2... \\ 
        in & AmazonBasics USB 2.0 ... & Is it compatible with Nintendo switch? -> Dono but working good nice product. &  AmazonBasics USB 2.0 ...& Is this compatible with MacOS? -> Yes. & 0& The answer to Product B's question does not provide information for A...
        \\
        uk &HDMI Media Player, Black Mini 1080p Full-HD Ultra...& Is it possible to power this through a usb cable? -> It has to be plugged in using the power lead... & MDN HD1080B 1080p Full-HD Ultra Portable Digital Media Player... & Can it be powered by a USB cable? I see on the pictures that power cable is USB on one end -> The USB port is for an external drive. & 2 & The question for both Product A and Product B pertains to the power source of the media players and whether they can be powered through a USB cable...\\

\bottomrule
    \end{tabular} 
    \caption{Examples of data samples in \OurDatarank. All the data is translated into English.}
    \label{tab:datasetexample3}
\end{table*}
\section{Case study}
\label{case_study}
Table~\ref{tab:casestudy} demonstrates four real cases concerning single/cross-market question answering. We see that the absence of useful information, such as related reviews or questions, within a single marketplace leads to inaccurate answers. For instance, in case 1, the retrieved reviews fail to provide sufficient information, resulting in a generated answer of ``I have no idea what to answer.'' In contrast, relevant and useful information is more likely to be available in the larger auxiliary marketplace. For instance, in case 4, the model successfully retrieves a similar question, ``Can you put an external microphone?'' from the 
\textbf{us} marketplace, aligning the answer more closely with the ground-truth answer.
\section{Dataset Examples}
\label{sec:dataset_examples}
We show some examples from \OurData to provide a more comprehensive view of our data. Table \ref{tab:datasetexample1} shows some examples from \OurData. For each example, we show the title of a product, a random review, and a question-answer pair of the product.

To provide a more comprehensive understanding of our dataset and task, we also show some examples of the GPT-4 annotated \OurDataans (Table \ref{tab:datasetexample2}) and \OurDatarank (Table \ref{tab:datasetexample3}), respectively.
\end{document}